\documentclass{ieeeaccess}
\usepackage{cite}
\usepackage{amsmath,amssymb,amsfonts}
\usepackage{algorithmic}
\usepackage{graphicx}
\usepackage{textcomp}
\usepackage{mathtools}
\usepackage{bbm}
\usepackage{url}
\def\BibTeX{{\rm B\kern-.05em{\sc i\kern-.025em b}\kern-.08em
    T\kern-.1667em\lower.7ex\hbox{E}\kern-.125emX}}
\begin{document}
\history{Received September 3, 2021, accpeted September 18, 2021. Date of publication xxxx 00, 0000, date of current version xxxx 00, 0000.}
\doi{10.1109/ACCESS.2021.3115476}

\title{Video Action Understanding}
\author{\uppercase{Matthew S. Hutchinson}
\uppercase{and Vijay N. Gadepally}, \IEEEmembership{Senior Member, IEEE} 
\address[]{Massachusetts Institute of Technology Lincoln Laboratory Supercomputing Center, Lexington, MA 02421 USA}
\tfootnote{Corresponding Author: Matthew S. Hutchinson (hutchinson@alum.mit.edu) \\
\medskip
This work was supported by the United States Air Force Research Laboratory and the United States Air Force Artificial Intelligence
Accelerator and was accomplished under Cooperative Agreement FA8750-19-2-1000.}}

\markboth
{M. S. Hutchinson and V. N. Gadepally: Video Action Understanding}
.

\begin{abstract}
    Many believe that the successes of deep learning on image understanding problems can be replicated in the realm of video understanding.  However, due to the scale and temporal nature of video, the span of video understanding problems and the set of proposed deep learning solutions is arguably wider and more diverse than those of their 2D image siblings.  Finding, identifying, and predicting actions are a few of the most salient tasks in this emerging and rapidly evolving field.  With a pedagogical emphasis, this tutorial introduces and systematizes fundamental topics, basic concepts, and notable examples in supervised video action understanding.  Specifically, we clarify a taxonomy of action problems, catalog and highlight video datasets, describe common video data preparation methods, present the building blocks of state-of-the-art deep learning model architectures, and formalize domain-specific metrics to baseline proposed solutions.  This tutorial is intended to be accessible to a general computer science audience and assumes a conceptual understanding of supervised learning.
\end{abstract}

\begin{keywords}
action detection, action localization, action prediction, action proposal, action recognition, action understanding, video understanding
\end{keywords}

\titlepgskip=-15pt

\maketitle

\section{Introduction}

\begin{figure*}
    \centering
    \includegraphics[width=1.0\textwidth]{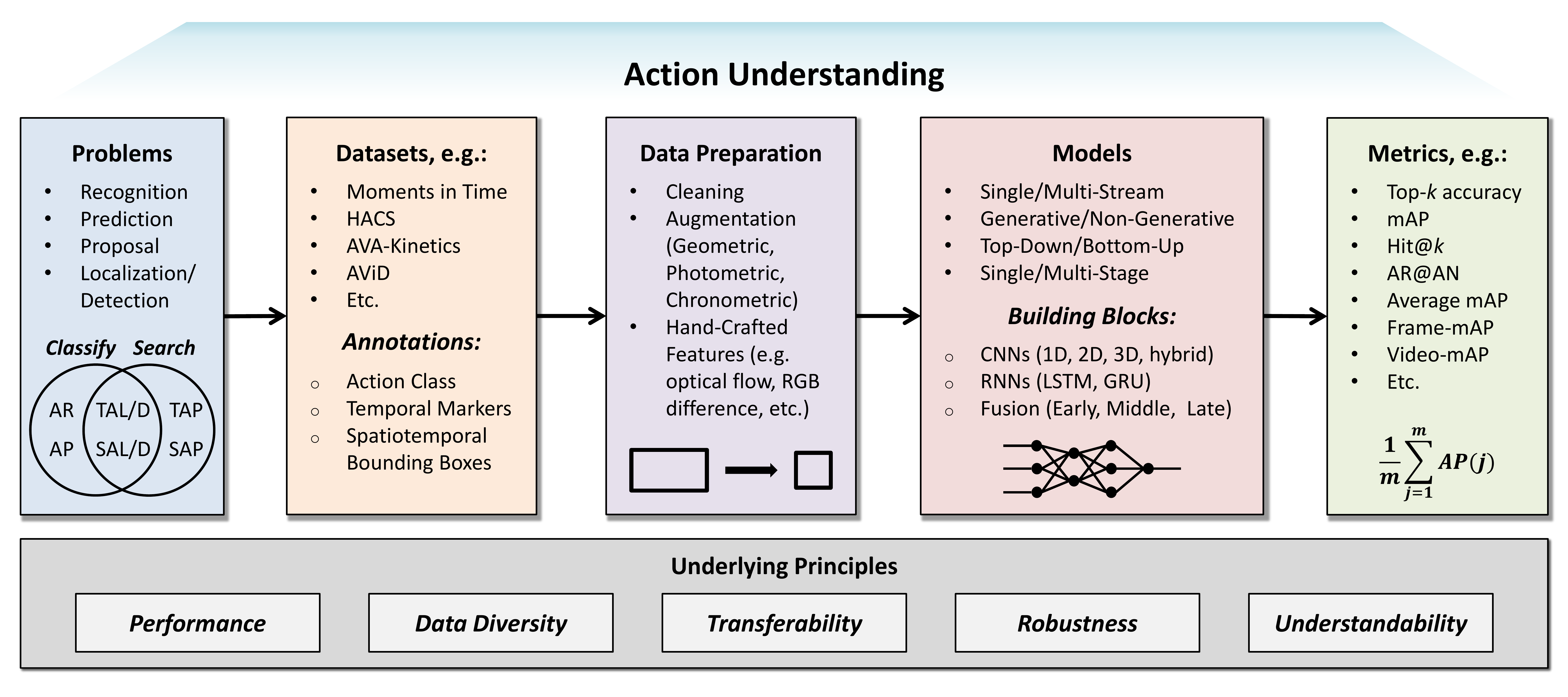}
    \caption{Overview of action understanding steps (problem formulation, dataset selection, data preparation, model development, and metric-based evaluation) and underlying principles (computational performance, data diversity, transferability, robustness, and understandability. This serves as the framework for this tutorial.}
    \label{fig:overall-figure}
\end{figure*}

Video understanding is a natural extension of deep learning research efforts in computer vision.  The image understanding field greatly benefited from the application of artificial neural network (ANN) machine learning (ML) methods.  Many image understanding problems—object recognition, scene classification, semantic segmentation, etc.—have workable deep learning “solutions.”  FixEfficientNet-L2 currently boasts 88.5\%/98.7\% Top-1/Top-5 accuracy on the ImageNet object classification task \cite{touvron2020fixing, ILSVRC15}.  Hikvision Model D scores 90.99\% Top-5 accuracy on the Places2 scene classification task \cite{ILSVRC15, 7968387}.  High-resolution network - object contextual representations (HRNet-OCR) with hierarchical multi-scale attention (HMS) yields a mean intersection over union (IoU) of 85.1\% on the Cityscapes semantic segmentation task \cite{Cordts_2016_CVPR, Borse_2021_CVPR}.  Naturally, many hope that deep learning methods can achieve similar levels of success on video understanding problems.  

Drawing from Diba et al. (2019) \cite{10.1007/978-3-030-58558-7_35}, \textit{semantic video understanding} is a combination of understanding the scene/environment, objects, actions, events, attributes, and concepts.  This article focuses on the action component and is presented as a tutorial that introduces a common set of terms and tools, explains basic and fundamental concepts, and provides concrete examples.  We intend this to be accessible to a general computer science audience and assume readers have a basic understanding of supervised learning---the paradigm of learning from input-output examples.

\subsection{Action Understanding}\label{Action-Understanding}

While the literature often uses the terms \textit{action} and \textit{activity} synonymously \cite{Ke_2013, CHAQUET2013633, cheng2015advances}, we prefer to use action in this article for a few reasons.  First, action is the dominant term used across the field, so we would need a significant reason to divert from that term.  Second, the use of activity is generally biased towards human actors rather than non-human actors and phenomenon. Examples of non-human actors and phenomenon performing actions include a dog running, a cloud floating, and a wheel turning.  We prefer action over activity for its broader human and non-human applicability. Third, activity recognition is a term already used in several non-video domains \cite{6208895, 10.1145/1964897.1964918, WANG20193}.  Meanwhile, action recognition is primarily a computer vision and video-based term.

But what is an action?  Kang and Wildes (2016) \cite{kang2016review} consider an action to be “a motion created by the human body, which may or may not be cyclic."  Zhu et al. (2016) \cite{ZHU201642} define action as an “intentional, purposive, conscious and subjectively meaningful activity."  Several human action surveys create a spectrum of action complexity from gestures to interactions or group activities \cite{ZHU201642, cheng2015advances, GUO20143343}. Unlike these surveys, we use a broader definition of action, one that includes actions of both human and non-human actors because: 1) video datasets are being introduced that use this broader definition \cite{8651343, monfort2019multimoments}; 2) most deep learning metrics and methods are equally applicable to both settings; and 3) the colloquial use of action has no distinction between human and non-human actors.  Merriam-Webster’s Dictionary and the Oxford English Dictionary define action as “an act done” and “something done or performed,” respectively \cite{merriam-webster-action, oed-action}.  Therefore, this article defines \textit{action} as something done or performed intentionally or unintentionally by a human or non-human actor from which a human observer could derive meaning.  This includes everything from low-level gestures and motions to high-level group interactions.

As shown in Fig. \ref{fig:overall-figure}, \textit{action understanding} encompasses action problems, video action datasets, data preparation techniques, deep learning models, and evaluation metrics. Underlying these steps are computer vision and supervised learning principles of computational performance, data diversity, transferability, model robustness, and understandability.

\subsection{Related Work and Our Contribution}

\begin{table}[t]
    \centering
    \caption{Coverage of surveys and tutorials on action understanding. Tabular information includes year of publication, action coverage: human (H) and/or non-human (N), topic coverage: datasets (Ds), metrics (Mc), models/methods (Md), and problem coverage: 1) action recognition; 2) action proposal; 3) temporal action proposal; 4) temporal action localization/detection; and 5) spatiotemporal action localization/detection. }
    \label{tab:related-work}
    \setlength{\tabcolsep}{2pt}
    \begin{tabular}{| l l | c c | c c c | c c c c c |}
        \hline
        & & \multicolumn{2}{c|}{Actions} & \multicolumn{3}{c|}{Topics} & \multicolumn{5}{c|}{Problems} \\
        Survey/Tutorial & Year & H & N & Ds & Mc & Md & 1 & 2 & 3 & 4 & 5 \\ %AR & AP & TAP & TAL & SAL \\
        \hline
        Poppe \cite{POPPE2010976} & 2010 & \checkmark & & \checkmark & & \checkmark & \checkmark & & & & \\
        Weinland et al. \cite{WEINLAND2011224} & 2011 & \checkmark & &\checkmark & & \checkmark & \checkmark & & & & \\
        Ahad et al. \cite{6060230} & 2011 & \checkmark & & \checkmark & & \checkmark & & & \checkmark & & \\
        Chaquet et al. \cite{CHAQUET2013633} & 2013 & \checkmark & &\checkmark & & \checkmark & \checkmark & & & & \\
        Guo and Lai \cite{GUO20143343} & 2014 & \checkmark & &\checkmark & & \checkmark & \checkmark & \checkmark & & & \\
        Cheng et al. \cite{cheng2015advances} & 2015 & \checkmark & &\checkmark & & \checkmark & \checkmark & \checkmark & & & \\
        Zhu et al. \cite{ZHU201642} & 2016 & \checkmark & & & & \checkmark & \checkmark & & & & \\
        Kang and Wildes \cite{kang2016review} & 2016 & \checkmark & & \checkmark & \checkmark & \checkmark & \checkmark & \checkmark & & \checkmark & \\
        Zhang et al. \cite{ZHANG201686} & 2016 & \checkmark & & \checkmark & & \checkmark & \checkmark & & & \checkmark & \\
        Herath et al. \cite{HERATH20174} & 2017 & \checkmark & & \checkmark & & \checkmark & \checkmark & & & & \\
        Koohzadi and Charkari \cite{koohzadi2017survey} & 2017 & \checkmark & & & & \checkmark & \checkmark & & & & \\
        Asadi-Aghbolaghi et al. \cite{7961779} & 2017 & \checkmark & & \checkmark & & \checkmark & \checkmark & & & & \\
        Kong and Fu \cite{kong2018human} & 2018 & \checkmark & & \checkmark & & \checkmark & \checkmark & \checkmark & & & \\
        Zhang et al. \cite{s19051005} & 2019 & \checkmark & & \checkmark & & \checkmark & \checkmark & \checkmark & & \checkmark & \\
        Bhoi \cite{bhoi2019spatiotemporal} & 2019 & \checkmark & \checkmark & \checkmark & & \checkmark & & & & & \checkmark \\
        Singh and Vishwakarma \cite{video-benchmarks} & 2019 & \checkmark & & \checkmark & & & \checkmark & & & \checkmark & \\
        Xia and Zhan \cite{9062498} & 2020 & \checkmark  & \checkmark & \checkmark & \checkmark & & & & \checkmark & \checkmark & \\
        Rasouli \cite{rasouli2020deep} & 2020 & \checkmark & \checkmark & \checkmark & \checkmark & \checkmark & & \checkmark & & & \\
        Yadav et al. \cite{YADAV2021106970} & 2021 & \checkmark & & \checkmark & & \checkmark & \checkmark & \checkmark & & \checkmark & \\
        \hline
        \textbf{Ours} & 2021 & \checkmark & \checkmark & \checkmark & \checkmark & \checkmark & \checkmark &\checkmark &\checkmark & \checkmark & \checkmark \\ 
        \hline
    \end{tabular}
\end{table}

Table \ref{tab:related-work} shows a selection of surveys and tutorials on action understanding written in the last decade.  Yet, few focus on more than one or two action problems or present more than a narrow coverage of datasets.  Additionally, the majority only consider a narrow (human) definition of actions and have little or no discussion of metrics. Of the more recent examples, Kong and Fu \cite{kong2018human}, Xia and Zhan \cite{9062498}, and Rasouli \cite{rasouli2020deep} are the most thorough in their independent directions.  We recommend these works for a more advanced analysis and comparison of deep learning models.  Relative to the literature noted above, we present this article as a tutorial to introduce and systematize a balance of topics across datasets, methods, and metrics rather than a deeply technical methods-heavy survey. To that goal, this tutorial contributes the following:

\begin{itemize}
    \item Clear definitions of recognition, prediction, proposal, and localization/detection video action problems.
    \item An extensive catalog of video action datasets and discussion of those most relevant to each action problem.
    \item Descriptions of the oft neglected, yet important methods of video data preparation and feature extraction.
    \item Explanations of common deep learning model building blocks with domain-specific examples.
    \item Groupings of state-of-the-art model architectures.
    \item Formal definitions of evaluation metrics across the span of video action problems.
\end{itemize}

This article is organized in the following way.  Section \ref{problems} defines and organizes action understanding problems.  Section \ref{datasets} catalogs video action datasets by annotation type which directly relates to the problems for which they are applicable. Section \ref{data_preparation} provides an introduction to video data and data preparation techniques.  Section \ref{models} presents basic model building blocks and organizes state-of-the-art methods into families.  Section \ref{metrics} defines standard metrics used across these problems, formally shows how they are calculated, and points to examples of their usage in high-profile action understanding competitions.  Section \ref{conclusion} summarizes and concludes the tutorial.  Note that because the breadth of topics covered is large, we chose to confine the scope of this tutorial to supervised learning as this is the dominant paradigm employed for video action understanding.

\section{Problems}\label{problems}

Several problems fall under the umbrella of action understanding. In this section, we introduce a taxonomy, provide definitions, and indicate disagreements in the literature.

\subsection{Taxonomy}\label{Taxonomy}

We organize six main action understanding problems into overlapping classification and search bins.  Classification involves labeling videos by their action class.  Search involves temporally or spatiotemporally finding action instances.

\begin{figure*}
    \centering
    \includegraphics[width=1.0\linewidth]{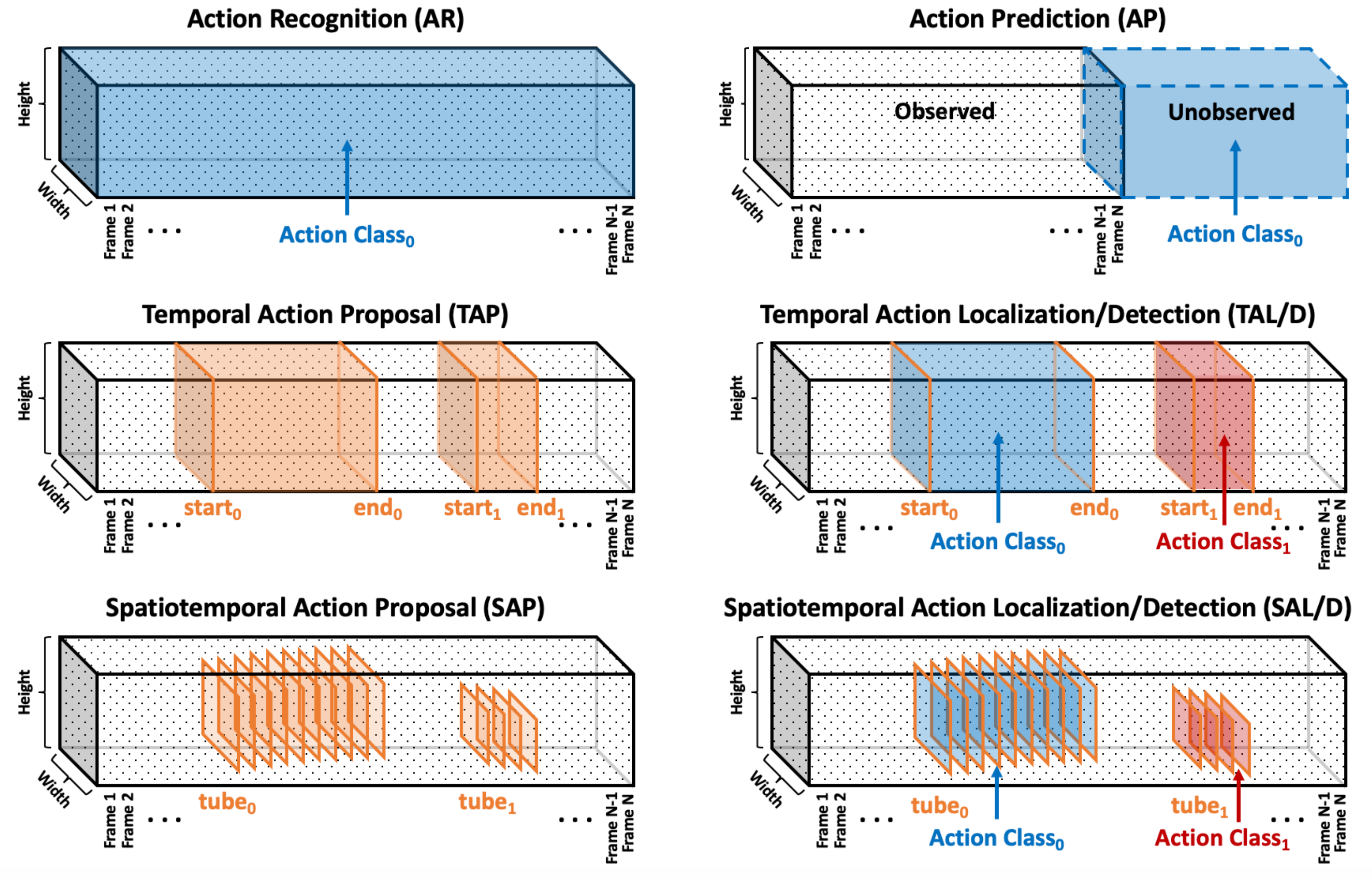}
    \caption{An overview of the main action understanding problems.  Video is depicted as a 3D volume where $N$ frames are densely stacked along a temporal dimension (left-to-right).  Action recognition (upper left) shows how an action class label is assigned to the entirety of the video or video clip.  Action prediction (upper right) shows how an action class label is assigned to a yet unobserved or only partial observed portion of a video.  Temporal action proposal (middle left) shows how temporal regions of likely action are bounded by start and end frames.  Temporal action localization/detection (middle right) shows how action class labels are assigned to temporal regions of likely action that are bounded by start and end frames.  Spatiotemporal action proposal (bottom left) shows how "tubes" or "tubelets" are formed from bounding boxes across frames indicating spatiotemporal regions of likely action.  Spatiotemporal action localization/detection (bottom right) shows how action class labels are assigned to spatiotemporal tubes of likely action.}
    %\Description{Videos are shown as long rectangular prisms.  Arrows point to colored regions indicating action classes, action proposals, and action detections.}
    \label{fig:action-problems}
\end{figure*}

\subsubsection{Definitions}

\textit{Action Recognition (AR)} is the process of classifying a complete input (either an entire video or a specified segment) by the action occurring in the input.  If the action instance spans the entire length of the input, then the problem is known as \textit{trimmed action recognition}.  If the action instance does not span the entire input, then the problem is known as \textit{untrimmed action recognition}.  Untrimmed action recognition is generally more challenging because a model needs to complete the action classification task while disregarding non-action background segments of the input.

\textit{Action Prediction (AP)} is the process of classifying an incomplete input by the action yet to be observed.  One sub-problem is \textit{action anticipation (AA)} in which no portion of the action is yet observed and classification is entirely based on observed contextual clues. Another is \textit{early action prediction (EAP)} in which a portion, but not the entirety, of the action instance is observed.  Both AR and AP are classification problems, but AP often requires a dataset with temporal annotations so that there is a clear delimiter between a "before-action" segment and "during-action" segment for AA or between "start-action" and "end-action" for EAP.  

\textit{Temporal Action Proposal (TAP)} is the process of partitioning an input video into segments (consecutive series of frames) of action and inaction by indicating start and end markers of each action instance.  \textit{Temporal Action Localization/Detection (TAL/D)} is the process of creating temporal action proposals and classifying each action.

\textit{Spatiotemporal Action Proposal (SAP)} is the process of partitioning an input video by both space (bounding boxes) and time (per-frame or start/end markers of a segment) between regions of action and inaction.  If a linking strategy is applied to bounding boxes across several frames, the regions of actions that are constrained in the spatial and temporal dimensions are often referred to as \textit{tubes} or \textit{tubelets}. \textit{Spatiotemporal Action Localization/Detection (SAL/D)} is the process of creating spatiotemporal action proposals and classifying each frame's bounding boxes (or action tubes when a linking strategy is applied).

\subsubsection{Literature Observations}

This taxonomy and these definitions are intended to clarify several term discrepancies in the literature.  First, recognition and classification are sometimes used interchangeably (e.g., \cite{4270162, 4270157, Girdhar_2017_CVPR}).  We believe that should be avoided because both recognition (an identification task) and prediction (an anticipation task) require arranging inputs into categories (i.e., classification).  To use recognition and classification synonymously incorrectly equates recognition and prediction.  Second, localization and detection are often used interchangeably (e.g., \cite{Shou_2016_CVPR, Zhao_2017_ICCV, Chao_2018_CVPR}).  However, in this case, because the task involves finding and identifying, we feel the terms are appropriate.  While detection appears slightly more prevalent in the temporal action literature and localization appears slightly more prevalent in the spatiotemporal action literature, this article will remain neutral and use localization/detection (L/D) together as a single term.  Third, action proposal and action proposal generation are used interchangeably (e.g., \cite{Lin_2018_ECCV, Gao_2018_ECCV, Liu_2019_CVPR}).  We chose to use the former because proposal can be defined as the act of generating a proposal.  Proposal generation is therefore redundant. An important takeaway is that the literature contains many examples where different terms refer to the same action problem (e.g., \cite{bhoi2019spatiotemporal} and \cite{ESCORCIA2020102886}).  Similarly, there are many examples where the same terms refer to different problems (e.g., \cite{Zeng_2019_ICCV} and \cite{Xu_2017_ICCV}).  To compound the issue, many video action datasets can be applied to more than one of these problems. We encourage readers to pay careful attention to these terms when venturing into this field.

Another notable observation from the literature is that while TAP and TAL/D are sometimes studied independently, SAP is not studied outside of a SAL/D framework.  Therefore, the remainder of this article does not refer to SAP independently of SAL/D.

\subsection{Related Problems}

Here, we define a few problems related to, but not included in, our main taxonomy and cite literature for further reading.

\textit{Action instance segmentation (AIS)} is the labeling of individual instances or examples of an action within the same video even when these action instances may overlap in both space and time. Therefore, AIS is a constraint that can be placed on top of TAL/D or SAL/D. For example, a model performing SAL/D on a video of a concert may identify the frames and bounding boxes sections where the audience is shown and label the proposed temporal segment with the action “clapping.”  Applying the AIS constraint would require the model to divide the bounding boxes into each individual clapping member of the audience and track these individual actions across time.  Useful action instance segmentation literature includes Weinland et al. (2011) \cite{WEINLAND2011224}, Saha et al. (2017) \cite{saha2017spatiotemporal}, Ji et al. (2018) \cite{Ji_2018_ECCV}, and Saha et al. (2020) \cite{Saha2020}.

\textit{Dense captioning} is the generation of sentence descriptions for videos.  This problem spans several of the video understanding semantic components and is worth noting because it is often paired with action understanding problems in public challenges \cite{ghanem2017activitynet, ghanem2018activitynet, activitynetchallenge2019, activitynetchallenge2020}.  Similarly, video captioning datasets (such as MSVD \cite{chen2011collecting}, MVAD \cite{torabi2015using}, MPII-MD \cite{Rohrbach_2015_CVPR} and ActivityNet Captions \cite{krishna2017dense}) will sometimes be included in video action understanding dataset lists.  For more on video captioning, Li et al. (2019) \cite{8627985} present a survey on methods, datasets, difficulties, and trends. 

\textit{Action spotting (AS)}, proposed by Alwassel et al. (2018) \cite{Alwassel_2018_ECCV}, is the process of finding any temporal occurrence of an action in a video while observing as little as possible.  This differs from TAL/D in two ways.  First, AS requires finding only a single frame within the action instance segment rather than start and end markers.  Second, AS is concerned with the efficiency of the search process.  

\textit{Object tracking} is the process of detecting objects and linking detections between frames to track them across time. Object tracking is a relevant related problem because some metrics used for object detection in videos were adopted in video action detection \cite{kpkl2019watch,pascal-voc}.  We recommend Yao et al. (2019) \cite{yao2020video} for a recent and broad survey on video object segmentation and tracking.

\section{Datasets}\label{datasets}

\begin{table*}[t]
    \scriptsize
    \centering
    \caption{Thirty historically influential, current state-of-the-art, and emerging benchmarks of video action datasets. Tabular information includes dataset name, year of publication, citations on Google Scholar as of May 2021, number of action classes, number of action instances, actors: human (H) and/or non-human (N), annotations: action class (C), temporal markers (T), spatiotemporal bounding boxes/masks (S), and theme/purpose.}
    \label{tab:datasets}
    \begin{tabular}{|l c r | r r | c c | c c c | l |}
        \hline
        % NOTE: only those uncommented have had their citations updated
         & & & \multicolumn{2}{c}{Action} & \multicolumn{2}{|c|}{Actors} & \multicolumn{3}{c|}{Annotations} & \\
         Video Dataset & Year & Cited & Classes & Instances & H & N & C & T & S & Theme/Purpose \\
        \hline
         KTH \cite{1334462} & 2004 & 4,246 & 6 & 2,391 & \checkmark & & \checkmark & & & B/W, static background \\
        %CAVIAR \cite{caviar} & 2004 & 49 & 9 & $>$28 & \checkmark & & \checkmark & & \checkmark & surveillance \\
         Weizmann \cite{1544882} & 2005 & 2,068 & 10 & 90 & \checkmark & & \checkmark & & & human motions \\
         %ViSOR \cite{4607676, 4730416} & 2005 & 47 & n/a & n/a & \checkmark & & \checkmark & & & surveillance \\
         %ETISEO \cite{4118811, 4425357} & 2005 & 183 & 15 & n/a & \checkmark & & \checkmark & & & human motions \\
         %IXMAS \cite{WEINLAND2006249} & 2006 & 977 & 13 & 390 & \checkmark & & \checkmark & & & B/W, partial occlusion \\
         %UCF Aerial \cite{ucfaerial} & 2007 & n/a & 9 & n/a & \checkmark & & \checkmark & & \checkmark & aerial-view \\
         %CASIA Action \cite{4270503, 4270101} & 2007 & 242 & 15 & 1,446 & \checkmark & & \checkmark & & & multi-view, outdoors \\
         Coffee \& Cigarettes \cite{4409105} & 2007 & 526 & 2 & 246 & \checkmark & & \checkmark & \checkmark & \checkmark & movies and TV  \\
         %UIUC Action \cite{10.1007/978-3-540-88682-2_42} & 2008 & 378 & 14 & 532 & \checkmark & & \checkmark & & & action repetition \\
         %UCF Sports \cite{4587727} & 2008 & 1,269 & 10 & 150 & \checkmark & & \checkmark & & \checkmark & sports \\
         %UCF ARG \cite{ucfarg} & 2008 & n/a & 10 & 480 & \checkmark & & \checkmark & & \checkmark & multi-view, aerial-view \\
         %Hollywood (HOHA) \cite{4587756} & 2008 & 3,727 & 8 & n/a & \checkmark & & \checkmark & & & movies \\
         %Cambridge-Gesture \cite{4547427} & 2008 & 298 & 9 & 900 & \checkmark & & \checkmark & & & gestures \\
         %BEHAVE \cite{blunsden2010behave} & 2009 & 134 & 10 & 163 & \checkmark & & \checkmark & & \checkmark & human-human interaction\\
         %URADL \cite{5459154} & 2009 & 574 & 10 & 150 & \checkmark & & \checkmark & & & daily activities \\
         %UCF11 \cite{5206744} & 2009 & 1,183 & 11 & 3,040 & \checkmark & & \checkmark & & & web videos \\
         %MSR-I \cite{5719621} & 2009 & 181 & 3 & n/a & \checkmark & & \checkmark & & \checkmark & activities \\
         %i3DPost MuHAVi \cite{5430066} & 2009 & 179 & 12 & $>$1,000 & \checkmark & & \checkmark & & & multi-view, studio \\
         Hollywood2 \cite{5206557} & 2009 & 1,488 & 12 & 3,669 & \checkmark & & \checkmark & & & movies \\
         VIRAT \cite{5995586} & 2011 & 634 & 23 & $\sim$10,000 & \checkmark & & \checkmark & \checkmark & \checkmark & surveillance, aerial-view \\
        HMDB51 \cite{6126543} & 2011 & 2,428 & 51 & $\sim$7,000 & \checkmark & & \checkmark & & & human motions \\
         %CAD-60 \cite{sung2012unstructured} & 2011 & 549 & 12 & 60 & \checkmark & & \checkmark & & \checkmark & RGB-D, daily activities \\
         %GTEA \cite{fathi2011learning, Li_2015_CVPR} & 2011 & 492 & 71 & 526 & \checkmark & & \checkmark & & & egocentric, kitchen \\
         %CCV \cite{ccv} & 2011 & 288 & *20 & 9,317 & \checkmark & & \checkmark & & & web videos \\
         %ChaLearn \cite{chalearn} & 2011 & n/a & 86 & 50,000 & \checkmark &  & \checkmark & & & RGB-D, gestures and motions \\
         %RGBD-HuDaAct \cite{6130379} & 2011 & 393 & 12 & 1,189 & \checkmark & & \checkmark & & & RGB-D, daily activities \\
         %NATOPS \cite{5771448} & 2011 & 111 & 24 & 400 & \checkmark & & \checkmark & & & aircraft hand signaling \\
         %GTEA Gaze \cite{fathi2012learning} & 2012 & 331 & 40 & 331 & \checkmark & & \checkmark & \checkmark & & egocentric, kitchen \\
         %GTEA Gaze+ \cite{fathi2012learning, Li_2015_CVPR} & 2012 & 165 & 44 & 1,958 & \checkmark & & \checkmark & \checkmark & & egocentric, kitchen \\
         %BIT-Interaction \cite{10.1007/978-3-642-33718-5_22} & 2012 & 109 & 8 & 400 & \checkmark & & \checkmark & & & human-human interaction \\
         %LIRIS \cite{WOLF201414} & 2012 & 60 & 10 & n/a & \checkmark & & \checkmark & & \checkmark & RGB-D, office environment \\
         %MSR-DailyActivity3D \cite{6247813} & 2012 & 1,339 & 16 & 320 & \checkmark & & \checkmark & & & RGB-D, gestures \\
         UCF101 \cite{soomro2012ucf101} & 2012 & 3,183 & 101 & 13,320 & \checkmark & & \checkmark & & & web videos, expand UCF50 \\
         THUMOS'13 \cite{THUMOS13, idrees2017thumos, soomro2012ucf101} & 2013 & 191 & *101 & 13,320 & \checkmark & & \checkmark & & \checkmark & web videos, extend UCF101 \\
         J-HMDB-21 \cite{Jhuang_2013_ICCV} & 2013 & 567 & 51 & 928 & \checkmark & & \checkmark & & \checkmark & re-annotate HMDB51 subset \\
         %Mivia \cite{10.1007/978-3-642-41190-8_47} & 2013 & 21 & 7 & 490 & \checkmark & & \checkmark & & & RGB-D, daily activities \\
         %IAS-lab \cite{MUNARO201342, munaro2013evaluation} & 2013 & 31 & 15 & 540 & \checkmark & & \checkmark & & & RGB-D, human motions \\
         %WorkoutSU-10 \cite{10.1007/978-3-642-39094-4_74} & 2013 & 66 & 10 & 1,200 & \checkmark & & \checkmark & & & RGB-D, group activities \\
         %50Salads \cite{10.1145/2493432.2493482} & 2013 & 177 & 17 & 966 & \checkmark & & \checkmark & \checkmark & & RGB-D, kitchen \\
         %UWA3D-I \cite{10.1007/978-3-319-10605-2_48} & 2014 & 141 & 30 & $\sim$900 & \checkmark & & \checkmark & & & RGB-D, multi-view \\
         %MANIAC \cite{AKSOY2015118} & 2014 & 43 & 8 & 120 & \checkmark & & \checkmark & \checkmark & & RGB-D, egocentric, manipulations \\
         %Breakfast Action \cite{Kuehne_2014_CVPR} & 2014 & 203 & 48 & 11,267 & \checkmark & & \checkmark & \checkmark & & kitchen \\
         %Northwester-UCLA \cite{Wang_2014_CVPR} & 2014 & 222 & 10 & 1,475 & \checkmark & & \checkmark & & & RGB-D, multi-view \\
         Sports-1M \cite{Karpathy_2014_CVPR} & 2014 & 5,667 & 487 & 1,000,000 & \checkmark & & \checkmark & & & multi-label, sports \\
         ActivityNet200 (v2.3) \cite{Heilbron_2015_CVPR} & 2016 & 1,118 & 200 & 23,064 & \checkmark & & \checkmark & \checkmark & & untrimmed web videos \\
         %YouTube-8M \cite{abuelhaija2016youtube8m} & 2016 & 607 & *n/a & n/a & \checkmark & & \checkmark & & & multi-label \\
         %Charades \cite{10.1007/978-3-319-46448-0_31} & 2016 & 343 & 157 & 66,500 & \checkmark & & \checkmark & \checkmark & & crowd-sourced, daily activities \\
         %NTU RGB-D \cite{Shahroudy_2016_CVPR} & 2016 & 792 & 60 & 56,880 & \checkmark & & \checkmark & & & RGB-D, multi-view \\
         % Micro-Videos \cite{nguyen2016open} & 2016 & 27 & *n/a & n/a & \checkmark & \checkmark & \checkmark & & & micro-videos (e.g., Vine, Tik-Tok) \\
         % JAAD \cite{rasouli2017they, rasouli2018role} & 2017 & 53 & n/a & 654 & \checkmark & & \checkmark & & \checkmark & pedestrians \\
         % DAHLIA \cite{7961782} & 2017 & 9 & 7 & 51 & \checkmark & & \checkmark & & & RGB-D, daily activities \\
         %PKU-MMD \cite{liu2017pku} & 2017 & 67 & 51 & 3,366 & \checkmark & & \checkmark & & & RGB-D, multi-view \\
         %SYSU 3DHOI \cite{Hu_2015_CVPR} & 2017 & 302 & 12 & 480 & \checkmark & & \checkmark & & & RGB-D, human-object inter. \\
         %DALY \cite{weinzaepfel2016human} & 2017 & 26 & 10 & 3,600 & \checkmark & & \checkmark & & \checkmark & daily activities \\
         %Okutama Action \cite{Barekatain_2017_CVPR_Workshops} & 2017 & 55 & 12 & ~4,700 & \checkmark & & \checkmark & & \checkmark & aerial view \\
         Kinetics-400 \cite{kay2017kinetics} & 2017 & 1,380 & 400 & 306,245 & \checkmark & & \checkmark & & & diverse web videos \\
         AVA \cite{Gu_2018_CVPR} & 2017 & 404 & 80 & $>$392,416 & \checkmark & & \checkmark &  & \checkmark & atomic visual actions \\
         %Something-Something \cite{Goyal_2017_ICCV} & 2017 & 182 & 174 & 108,499 & \checkmark & & \checkmark & & & human-object interactions \\
         %SLAC \cite{slac} & 2017 & 19 & 200 & $\sim$1,750,000 & \checkmark & & \checkmark & \checkmark & & sparse-labelled web videos \\
         Moments in Time (MiT) \cite{8651343} & 2017 & 212 & 339 & 836,144 & \checkmark & \checkmark & \checkmark & & & intra-class variation, web videos \\
         MultiTHUMOS \cite{multithumos} & 2017 & 305 & 65 & $\sim$16,000 & \checkmark & & \checkmark & \checkmark & & multi-label, extends THUMOS \\
         %VIENA$^2$ \cite{10.1007/978-3-030-20887-5_28} & 2018 & 7 & 25 & 15,000 & \checkmark & \checkmark & \checkmark & \checkmark & & pedestrians and vehicles \\
         %PRAXIS Gesture \cite{negin2018praxis} & 2018 & 16 & 29 & $\sim$4,600 & \checkmark & & \checkmark & & & RGB-D, gestures \\
         %UAV-GESTURE \cite{Perera_2018_ECCV_Workshops} & 2018 & 10 & 13 & 119 & \checkmark & & \checkmark & & \checkmark & aerial-view, gestures \\
         %Diving48 \cite{Li_2018_ECCV-diving} & 2018 & 25 & 48 & 18,404 & \checkmark & & \checkmark & & & diving motions (sports) \\
         %EPIC-KITCHENS-55 \cite{Damen_2018_ECCV} & 2018 & 209 & 125 & 39,594 & \checkmark & & \checkmark & \checkmark & \checkmark & egocentric, kitchen \\
         %YouCook2 \cite{zhou2018towards} & 2018 & 96 & n/a & $\sim$15,400 & \checkmark & & \checkmark & \checkmark & & web videos, kitchen \\
         Kinetics-600 \cite{carreira2018short} & 2018 & 115 & 600 & 495,547 & \checkmark & & \checkmark & & & extends Kinetics-400 \\
         %VLOG \cite{Fouhey_2018_CVPR} & 2018 & 41 & 30 & $\sim$122,000 & \checkmark & & \checkmark & \checkmark & & web videos, human-object inter. \\
         EGTEA Gaze+ \cite{Li_2018_ECCV} & 2018 & 94 & 106 & 10,325 & \checkmark & & \checkmark & \checkmark & \checkmark & egocentric, kitchen \\
         Something-Something-v2 \cite{mahdisoltani2018effectiveness} & 2018 & 12 & 174 & 220,847 & \checkmark & & \checkmark & & & extends Something-Something \\
         Charades-Ego \cite{sigurdsson2018charadesego} & 2018 & 39 & 157 & 68,536 & \checkmark & & \checkmark & \checkmark & & egocentric, daily activities \\
         %Youtube-8M Segments \cite{abuelhaija2016youtube8m} & 2019 & n/a & *** & n/a & \checkmark & & \checkmark & \checkmark & & multi-label, extends YouTube-8M \\
         Jester \cite{Materzynska_2019_ICCV} & 2019 & 37 & 27 & 148,092 & \checkmark & & \checkmark & & & crowd-sourced, gestures \\
         %LSVV-HRI \cite{ji2019largescale} & 2019 & 4 & 83 & 25,600 & \checkmark & & \checkmark & & & RGB-D, human-robot interaction \\
         %PIE \cite{Rasouli_2019_ICCV} & 2019 & 10 & 6 & $\sim$1,800 & \checkmark & & \checkmark & & \checkmark & pedestrians \\
         Kinetics-700 \cite{carreira2019short} & 2019 & 96 & 700 & $\sim$650,000 & \checkmark & & \checkmark & & & extends Kinetics-600 \\
         Multi-MiT \cite{monfort2019multimoments} & 2019 & 10 & 313 & $\sim$1,020,000 & \checkmark & \checkmark & \checkmark & & & multi-label, extends MiT \\
         HACS Clips \cite{Zhao_2019_ICCV} & 2019 & 64 & 200 & $\sim$1,500,000 & \checkmark & & \checkmark & & & trimmed web videos \\
         HACS Segments \cite{Zhao_2019_ICCV} & 2019 & 64 & 200 & $\sim$139,000 & \checkmark & & \checkmark & \checkmark & & extends and improves SLAC \\
         NTU RGB-D 120 \cite{8713892} & 2019 & 168 & 120 & 114,480 & \checkmark & & \checkmark & & & extends NTU RGB-D 60 \\ 
         EPIC-KITCHENS-100 \cite{damen2020rescaling} & 2020 & 8 & 97 & $\sim$90,000 & \checkmark & & \checkmark & \checkmark & \checkmark & extends EPIC-KITCHENS-55 \\
         AVA-Kinetics \cite{li2020avakinetics} & 2020 & 17 & 80 & $>$238,000 & \checkmark & & \checkmark &  & \checkmark & adds annotations, AVA+Kinetics \\
         %ARID \cite{xu2020arid} & 2020 & 0 & 11 & 3,784 & \checkmark & & \checkmark & & & dark (low-lighting) videos \\
         AViD \cite{piergiovanni2020avid} & 2020 & 5 & 887 & $\sim$450,000 & \checkmark & \checkmark & \checkmark & & & diverse peoples, anonymized faces \\
         FineGym \cite{Shao_2020_CVPR} & 2020 & 25 & 10 & 4,883 & \checkmark & & \checkmark & \checkmark & & gymnastics w/ sub-actions \\ 
         %TinyVIRAT \cite{demir2020tinyvirat} & 2020 & 1 & 26 & 12,829 & \checkmark & & \checkmark & & & low-resolution videos \\
         HAA500 \cite{chung2020haa500} & 2020 & 2 & 500 & $\sim$10,000 & \checkmark & & \checkmark & \checkmark & & course-grained atomic actions \\
        \hline
        \multicolumn{11}{l}{*Only 24 classes have spatiotemporal annotations. This subset is also known as UCF101-24.}
    \end{tabular}
\end{table*}

Data is critical to successful machine learning models. In this section, we catalog video action datasets, describe the diversity of foundational and emerging benchmarks, and highlight competitions using these datasets that are the pinnacle drivers of model development and progress.

\subsection{Video Action Dataset Catalog} \label{dataset-catalog}

The last two decades saw huge growth in available video action datasets.  To the best of our knowledge, we organized the most comprehensive collection of these datasets. We catalog over 130 video action datasets sorted by release year.  Due to the scale of this catalog, Table \ref{tab:datasets} shows a selection of thirty datasets that are historically influential, current state-of-the-art, or emerging benchmarks.  The extended catalog can be found in our online repository.\footnote{\label{online-repo}\url{https://github.com/hutch-matt/vau-tutorial/blob/master/VideoActionDatasetCatalog-updated.pdf}}

\subsubsection{Criteria}

To be included in our catalog, a dataset must meet each of the following criteria:
\begin{enumerate}
    \item Released between 2004 and 2020.
    \item Comprised of single-channel or multi-channel videos.
    \item Includes full-video or video segment annotations.
    \item Captures at least two action classes.
    \item Utilizes at least one of the following types of annotations: (C) action class labels, (T) temporal start/end segment markers or frame-level labels, or (S) spatiotemporal frame-level bounding boxes or masks.
\end{enumerate}

\subsubsection{Trends}

Several trends emerge from this catalog. 
First, these datasets grew considerably over the past two decades in both number of action classes and number of action instances.  This trend is present across all of the use cases and occurred over several orders of magnitude.  Larger datasets are essential for training deep learning models with often millions of parameters.  Second, datasets useful only for classification (mainly AR) are considerably larger and more prevalent than temporally or spatiotemporally annotated datasets.  This is expected because temporal markers or spatiotemporal bounding boxes are more challenging to create.  An annotator may require only a few seconds to identify whether a particular video contains a given action but would need much more time to mark the start and end of an action.  Additionally, solving AR is often considered a prerequisite for effective TAL/D or SAL/D.  Therefore, recognition research generally precedes localization/detection research.     

\subsection{Foundational and Emerging Benchmarks}

We organize datasets in three groups, those with: 
\begin{enumerate}
    \item only action class annotations primarily for AR;
    \item temporal annotations most useful for TAP, TAL/D, and sometimes AP; and
    \item spatiotemporal annotations most useful for SAL/D.
\end{enumerate}
Because many of the earlier influential video action datasets such as KTH and Weizmann are described at length in previous survey papers \cite{6060230, CHAQUET2013633, kong2018human}, we focus on the current largest and highest quality datasets.  

\subsubsection{Action Recognition Datasets} \label{action recognition datasets}

\begin{figure*}
    \centering
    \includegraphics[width=1.0\linewidth]{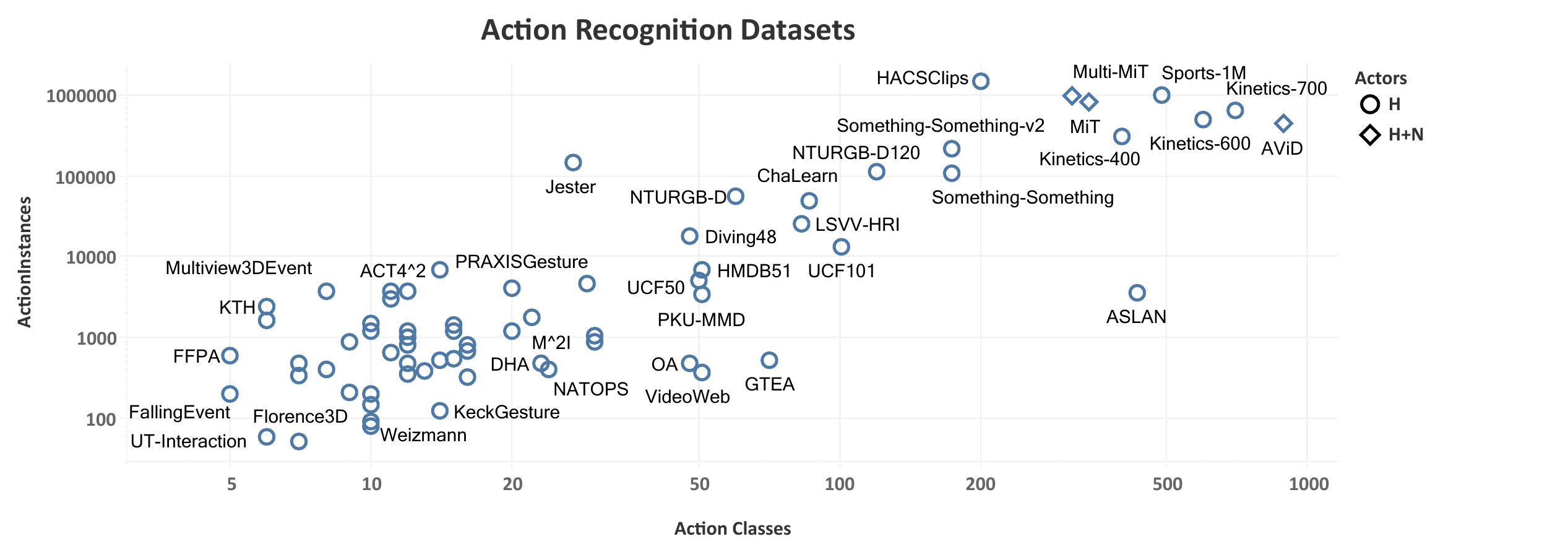}
    \caption{Datasets with only action class annotations mainly useful for action recognition (AR). Note that the plot is log-scaled in both dimensions.  Datasets in the upper right (e.g., Kinetics-700, Sport-1M, AViD) have the greatest number of action classes as well as action instances.  Many of the earlier influential datasets were much smaller and can be found in the lower left (e.g., KTH and Weizmann).  Because space on the plot is limited, not all AR datasets are labeled.  We recommend the catalog described in Section \ref{dataset-catalog} for more details.}
    \label{fig:AR-datasets}
\end{figure*}

Table \ref{fig:AR-datasets} plots AR-focused datasets by number of classes and number of instances.  Here we describe some of the largest and highest quality among them.

\textit{Sports-1M} \cite{Karpathy_2014_CVPR} was produced as a benchmark for comparing convolutional neural networks (CNNs).  Examples of the 487 sports action classes include "cycling," "snowboarding," and "american football."  Note that some inter-class variation is low (e.g., classes include 23 types of billiards, 6 types of bowling, and 7 types of American football). Videos were collected from YouTube and weakly annotated using text metadata.  The one million videos are divided with a 70/20/10 training/validation/test split. On average, videos are $\sim$5.5 minutes long, and about 5\% are annotated with $>1$ class.  As one of the first large-scale datasets, Sports-1M was critical for demonstrating the effectiveness of CNN architectures for feature learning.

\textit{Something-Something} \cite{Goyal_2017_ICCV} (a.k.a. 20BN-SOMETHING-SOMETHING) was produced as a human-object interaction benchmark.  Examples of the 174 classes include "holding something," "turning something upside down," and "folding something."   Video creation was crowd-sourced through Amazon Mechanical Turk (AMT).  108,499 videos are divided with an 80/10/10 training/validation/test split.  Each single-instance lasts for 2--6 seconds. A second and larger version \cite{mahdisoltani2018effectiveness} was released in 2018.  It also added object annotations, reducing label noise, and improving video resolution.  These are important benchmarks for human-object interaction due to their scale and quality.

The \textit{Kinetics} dataset family was produced as "a large-scale, high quality dataset of URL links" to human action video clips focusing on human-object and human-human interactions.  Class examples from \textit{Kinetics-400} \cite{kay2017kinetics} include "hugging," "mowing lawn," and "washing dishes."  Clips were collected from YouTube and annotated by AMT crowd-workers.  The dataset consists of 306,245 videos, and within each class, 50 and 100 are reserved for validation and testing, respectively.  Each single-instance video lasts for $\sim$10 seconds.  Additional videos and classes were added in 2018 \cite{carreira2018short} and 2019 \cite{carreira2019short}.  These are among the most cited human action datasets in the field and continue to serve as a standard benchmark and pretraining source.

\textit{NTU RGB-D}\cite{Shahroudy_2016_CVPR} was produced for RGB-D human action recognition.  The multi-modal nature provides depth maps, 3D skeletons, and infrared in addition to RGB video.  Examples of the 60 human actions include "put on headphone," "toss a coin," and "eat meal."  Videos were captured with a Microsoft Kinect v2 in a variety of settings.  The dataset consists of 56,880 single-instance video clips from 40 different subjects in 80 different views. Training and validation splits are not specified.  It was improved in 2019 \cite{8713892} with additional classes, videos, and views.  This serves as a state-of-the-art benchmark for human AR with non-RGB modalities.

\textit{Moments in Time (MiT)} \cite{8651343} was produced  with a focus on broadening action understanding to include people, objects, animals, and natural phenomenon.  Examples of the 339 human and non-human action classes include "running," "opening," and "picking."  Clips were collected from a variety of internet sources and crowd-annotated by AMT.  Nearly one million single-instance 3-second videos are divided with a roughly 89/4/7 training/validation/test split. The dataset was improved to \textit{Multi-Moments in Time (M-MiT)} \cite{monfort2019multimoments} in 2019, increasing the number of videos, pruning vague classes, and increasing the number of labels per video (2.01 million total labels).  MiT and M-MiT are interesting benchmarks because of the focus on inter-class and intra-class variation.

\textit{Jester} \cite{Materzynska_2019_ICCV} (a.k.a. 20BN-JESTER) was produced as "a large collection of densely labeled video clips that show humans performing pre-defined hand gestures in front of laptop camera or webcam."  Examples of the 27 human hand gestures include "drumming fingers," "shaking hand," and "swiping down."  Data creation was crowd-sourced through AMT and organized with a 80/10/10 training/validation/test split.  Each single-instance video lasts for $\sim$3 seconds.  The Jester dataset is the first large-scale, semantically low-level human AR dataset.

\textit{Anonymized Videos from Diverse countries (AViD)} \cite{piergiovanni2020avid} uniquely 1) avoids western bias by providing human actions (and some non-human actions) from a diverse set of people and cultures; 2) anonymizes all faces to protect privacy; and 3) ensures that all videos are static with a creative commons license.  Most of the 887 classes are drawn from Kinetics \cite{carreira2019short}, Charades \cite{10.1007/978-3-319-46448-0_31}, and MiT \cite{8651343} while removing duplicates and any actions that involve the face (e.g., "smiling"). 159 actions not found in any of those datasets are also added.  Web videos in 22 different languages were annotated by AMT crowd-workers.  Approximately 450,000 videos were organized with a 90/10 training/validation split.  Each single-instance video lasts between 3 and 15 seconds.  We believe AViD will quickly become a foundational benchmark because of the emphasis on diversity of actors and privacy.

\subsubsection{Temporally Annotated Datasets} \label{temporally annotated datasets}

\begin{figure*}
    \centering
    \includegraphics[width=1.0\linewidth]{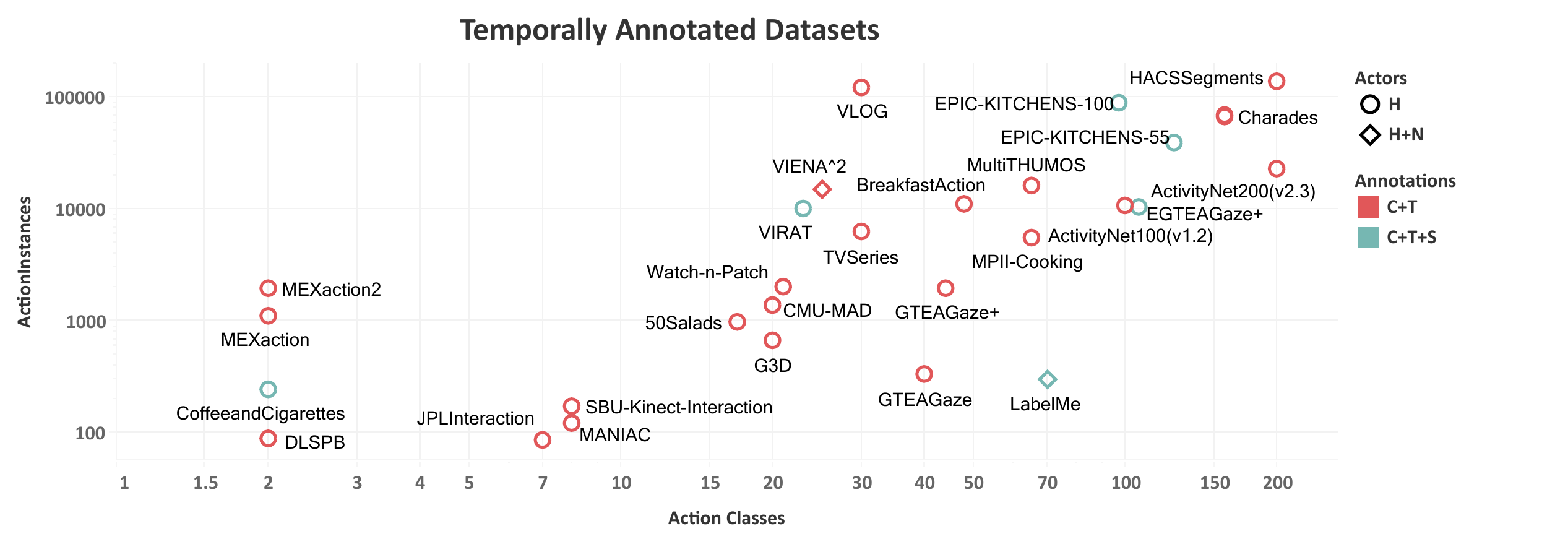}
    \caption{Datasets with temporal annotations useful for temporal action proposal (TAP), temporal action localization/detection (TAL/D), and possibly action prediction (AP). Note that the plot is log-scaled in both action instances and action classes dimensions.  The largest datasets with the greatest number of action classes and action instances (e.g., HACS Segments, Charades, and EPIC-Kitchens-100) can be found in the upper right. Even these "large" temporally annotated datasets are an order of magnitude smaller in both classes and instances than the largest action recognition datasets.  Also to note, the SLAC dataset \cite{slac} is excluded from this plot because while it has a very large number of temporally annotated action instances, the dataset was of poor quality.  HACS Segments was developed out of SLAC and has significantly fewer temporally annotated action instances.}
    %\Description{A scatter plot with class-only annotations datasets plotted along action instances on the y-axis and action classes on the x-axis. Most fall above 20 classes and 1000 actions. None have more than 200 classes and approximately 100,000 instances.}
    \label{fig:T-datasets}
\end{figure*}

Table \ref{fig:T-datasets} plots temporally annotated datasets by number of classes and action instances.  Here we describe some of the largest and highest quality among them.

The \textit{ActivityNet} dataset \cite{heilbron2014collecting, Heilbron_2015_CVPR} family was produced for both action recognition and detection. Example human action classes include "Drinking coffee," "Getting a tattoo," and "Ironing clothes."  \textit{ActivityNet 100 (v1.2)} is a 100-class dataset  divided into a 4,819 videos (7,151 instances) training set, a 2,383 videos (3,582 instances) validation set, and a 2,480 videos test set.  It was expanded to \textit{ActivityNet 200 (v1.3)} with 200-classes divided into a 10,024 videos (15,410 instances) training set, a 4,926 videos (7,654 instances) validation set, and a 5,044 videos test set.  On average, action instances are 51.4 seconds long. Web videos were temporal annotated by AMT crowd-workers.  ActivityNet remains as a foundational benchmark for TAP and TAL/D because of the dataset scope and size.  It is also commonly applied as an untrimmed multi-label AR benchmark.

\textit{Charades} \cite{10.1007/978-3-319-46448-0_31} was produced as a crowd-sourced dataset of daily human activities (e.g., "pouring into cup," "running," and "folding towel"). The dataset consists of 9,848 videos (66,500 temporal action annotations) with a roughly 80/20 training/validation split.  Videos were filmed in 267 homes with an average length of 30.1 seconds and an average of 6.8 actions per video.  Action instances average 12.8 seconds long.  \textit{Charades-Ego} used similar methodologies and the same 157 classes.  However, in this dataset, an egocentric (first-person) view and a third-person view is available for each video.  The dataset consists of 7,860 videos (68.8 hours) capturing 68,536 temporally annotated action instances.  Charades serves as a TAL/D benchmark along with ActivityNet, but it also useful as a multi-label AR benchmark because of the high average number of actions per video.  Charades-Ego presents a multi-view quality unique among large-scale daily human action datasets.    

\textit{MultiTHUMOS} \cite{multithumos} was produced as an extension of the dataset used in the 2014 THUMOS Challenge \cite{THUMOS14}.  Examples of the 65 human action classes include "throw," "hug," and "talkToCamera."  The 413 video (30 hours) dataset has 38,690 multi-label, frame-level annotations (an average of 1.5 per frame).  The total number of action instances---where an instance is a set of sequential frames with the same action annotation---is not reported.  The number of action instances per class is extremely variable ranging from "VolleyballSet" with 15 to "Run" with 3,500.  Each action instance lasts on average for 3.3 seconds with some lasting only 66 milliseconds (2 frames).  Like Charades, the MultiTHUMOS dataset offers a benchmark for multi-label TAP and TAL/D.  It stands out due to its dense multi-labeling scheme.

\textit{VLOG} \cite{Fouhey_2018_CVPR} was produced as an implicitly gathered large-scale daily human actions dataset.  Unlike previous daily human action datasets \cite{doi:10.1177/0278364913478446, 10.1007/978-3-319-46448-0_31, Goyal_2017_ICCV} in which the videos were created, VLOG was compiled from internet daily lifestyle video blogs (vlogs) and annotated by crowd-workers.  The method improves diversity of participants and scenes.  The dataset consists of 144,000 videos (14 days, 8 hours) using a 50/25/25 training/validation/test split.  The 30 classes are the objects with which the person is interacting (e.g., "Bag," "Laptop," and "Toothbrush").  Clips are labeled with these hand/object classes and temporally annotated with the state (positive/negative) of hand-object contact.  Because of the collection and annotation methods, VLOG brings actions in daily life datasets closer on par with other temporally annotated large-scale datasets.

\textit{HACS Segments} \cite{Zhao_2019_ICCV} was produced as a larger AR and TAL/D web-video dataset. Both HACS Segments and \textit{HACS Clips} (the AR portion) are improvements on the \textit{SLAC} dataset produced in the 2017 \cite{slac}.  HACS uses the same 200 human action classes as ActivityNet 200 (1.3).  Videos were collected from YouTube and temporally annotated by crowd-workers.  In HACS Segments 50,000 videos are divided with a 76/12/12 training/validation/test split.  The dataset contains 139,000 action instances (referred to as segments). Compared to ActivityNet, the number of action instances per video is greater (2.8 versus 1.5), and the average action instance duration is shorter (40.6 versus 51.4). HACS Segments is an emerging benchmark and provides a more challenging task for human TAP and TAL/D.

\subsubsection{Spatiotemporally Annotated Datasets} \label{spatiotemporally annotated datasets}

\begin{figure*}
    \centering
    \includegraphics[width=1.0\linewidth]{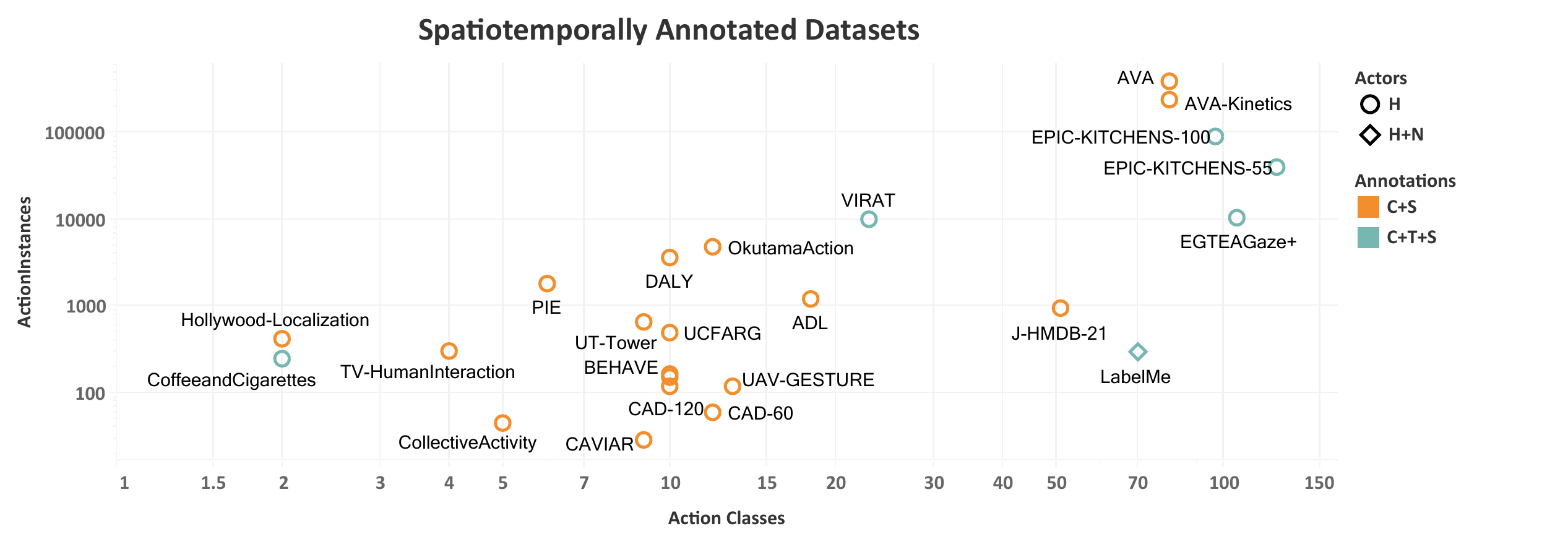}
    \caption{Datasets with spatiotemporal annotations useful for spatiotemporal action proposal (SAP) and spatiotemporal action localization/detection (SAL/D).  Those with class labels and only spatial annotations have actions which span the entirety of the video or video clip. Note the the plot is log-scaled in both instances and classes dimensions.  The largest of these datasets can be found in the upper right (e.g., AVA, AVA-Kinetics, and EPIC-KITCHENS-100).  Even these "large" spatiotemporally annotated datasets are an order of magnitude smaller than the largest action recognition datasets.}
    \label{fig:S-datasets}
\end{figure*}

Table \ref{fig:S-datasets} plots spatiotemporally annotated datasets.  Here we describe some of the largest and highest quality among them.  We also describe two smaller but still highly relevant datasets: UCF101-24 and J-HMDB-21.

\textit{VIRAT} \cite{5995586} was created as "a new large-scale surveillance video dataset designed to assess the performance of event recognition algorithms in realistic scenes."  It includes both ground and aerial surveillance videos.  Examples of the 23 classes include "picking up," "getting in a vehicle," and "exiting a facility."  The dataset consists of 17 videos (29 hours) with between 10 and 1,500 action instances per class.  Due to the camera distance across varying views, the human to video height ratio is between 2\% and 20\%.  Crowd-workers created bounding boxes around moving objects and temporal event annotations. While this is a smaller dataset, VIRAT is the highest quality surveillance-based spatiotemporal dataset and is used in the latest SAL/D competitions \cite{activitynetchallenge2019, activitynetchallenge2020}.  

\textit{UCF101-24}, the spatiotemporally labelled data subset of \textit{THUMOS'13} \cite{THUMOS13}, was produced as part of the THUMOS'13 challenge.  Examples of the 24 human action classes include "BasketballDunk," "IceDancing," "Surfing," and "WalkingWithDog."  Note, that the majority of the classes are sports.  It consists of 3,207 videos from the original UCF101 dataset \cite{soomro2012ucf101}.  Each video contains one or more spatiotemporally annotated action instances.  While multiple instances within a video have separate spatial and temporal boundaries, they have the same action class label.  Videos average $\sim$7 seconds long. The dataset is organized into three train/test splits.  While a small dataset, UCF101-24 remains a foundational benchmark for SAL/D.  

\textit{J-HMDB-21} \cite{Jhuang_2013_ICCV} was produced for pose-based action recognition.  Examples of the 21 human action classes include "brush hair," "climb stairs," and "shoot bow."  The dataset consists of 928 videos from the original HMDB51 dataset \cite{6126543} and is divided into three 70/30 train/test splits similar to UCF101.  Each video contains one action instance that lasts for the entire duration of the video.  2D joint masks and human-background segmentations were created by crowd-workers.  Because all of the action classes are human actions, bounding boxes could easily be derived from the joint masks or segmentation masks.  Along with UCF101-24, J-HMDB serves as a early foundational benchmark for SAL/D.

\textit{EPIC-KITCHENS-55} \cite{Damen_2018_ECCV} was produced as a large-scale benchmark for egocentric kitchen activities.  Examples of 149 human action classes include "put," "open," "pour," and "peel."  Videos were captured by head-mounted GoPro cameras on 32 individuals in 4 cities who were instructed to film anytime they entered their kitchen.  AMT crowd-workers located relevant actions and objects as well as created final action segment start/end annotations and object bounding boxes. The 432 videos (55 hours) are divided into a 272 video train/validation set, 106 video test set 1 (for previously seen kitchens), and a 54 video test set 2 (for previously unseen kitchens).  These correspond to 28,561, 8,064, and 2,939 action instances, respectively.  The dataset was improved to \textit{EPIC-KITCHENS-100} \cite{damen2020rescaling} increasing the number of videos, action instances, participants, and environments.  Annotation quality was also improved.  This dataset serves as a state-of-the-art egocentric kitchen activities benchmark.

\textit{Atomic Visual Actions (AVA)} \cite{Gu_2018_CVPR} was produced as the first large-scale spatiotemporally annotated diverse human action dataset.  Examples of the 80 classes include "swim," "write," and "drive."  The dataset consists of 437 15-minute videos with an approximately 55/15/30 training/validation/test split.  When only using the 60 most prominent classes (i.e., excluding those with fewer than 25 action instances), the dataset contains 214,622 training, 57,472 validation, and 120,322 test action instances.   Videos were gathered from YouTube and segments were annotated by crowd-workers.  Ground truth "tracklets" were calculated between manually annotated sections.  Because of the dataset scale, AVA serves as a large-scale multi-label benchmark for TAL/D.     

The \textit{AVA-Kinetics} dataset \cite{li2020avakinetics} was produced by using an existing large-scale human action recognition dataset and a spatiotemporal atomic action annotation schema.  The dataset combines a subset of videos from Kinetics-700 \cite{carreira2019short} and all videos from AVA \cite{Gu_2018_CVPR} for a total of 238,906 videos with a roughly 59/14/27 training/validation/test split.  For each 10-second video from Kinetics, a combination of algorithm and human crowd-workers created a bounding box for the frame with the highest person detection.  Crowd-workers then labeled the set of action instances performed by the person using the 80 possible action classes from the AVA dataset.  This dataset is an emerging benchmark because it improves upon AVA by dramatically expanding the number of annotated frames and increases the visual diversity.

\subsection{Competitions} \label{competitions}

\begin{table*}
    \footnotesize
    \centering
    \caption{Prominent Video Action Understanding Challenges 2013-2021.}
    \setlength{\tabcolsep}{5pt}
    \begin{tabular}{| l c l l l l l l |}
        \hline
         Workshop & Year & Conf. & Problem & Dataset & Metric(s) & Top Result & \#Teams \\
         \hline
         % https://www.crcv.ucf.edu/ICCV13-Action-Workshop/
         THUMOS \cite{THUMOS13} & 2013 & ICCV & AR & UCF101 & average accuracy & 85.90 \cite{Wang2013LEARINRIASF} & 17 \\
          & & & SAL/D & UCF101-24 & ROC AUC sIoU@0.2 & n/a & n/a \\
          \hline
         % https://www.crcv.ucf.edu/THUMOS14/submission.html
         THUMOS \cite{THUMOS14} & 2014 & ECCV & AR & UCF101+ & mAP & 0.71 \cite{Jain2014UniversityOA} & 11 \\
          & & & TAL/D & UCF101-20 & mAP tIoU@\{0.1,0.2,0.3,0.4,0.5\} & .4,.3,.3,.2,.1 \cite{Oneata2014TheLS} & 3 \\
          \hline
         % http://www.thumos.info/
         % https://arxiv.org/abs/1604.06182v1
         THUMOS \cite{THUMOS15, idrees2017thumos} & 2015 & CVPR & AR & UCF101+1 & mAP & 0.74 \cite{uts2016} & 11 \\
          & & & TAL/D & UCF101-20 & mAP tIoU@\{0.1,0.2,0.3,0.4,0.5\} & .4,.4,.3,.2,.2 \cite{Yuan2015ADSCSA} & 1 \\
          \hline
         % http://activity-net.org/challenges/2016/index.html
         % http://activity-net.org/challenges/2016/data/anet_challenge_summary.pdf
         ActivityNet \cite{Heilbron_2015_CVPR} & 2016 & CVPR & AR & ActivityNet 1.3 & mAP, Top-1 accuracy, Top-3 accuracy & 93.2, 88.1 \cite{xiong2016cuhk} & 26 \\
          & & & TAL/D & ActivityNet 1.3 & mAP-50, mAP-75, average-mAP & 42.5 & 6 \\
          \hline
         % http://activity-net.org/challenges/2017/index.html
         % https://arxiv.org/abs/1710.08011
         ActivityNet \cite{ghanem2017activitynet} & 2017 & CVPR & AR & ActivityNet 1.3 & Top-1 error & 8.8 \cite{ghanem2017activitynet} & n/a \\
          & & & AR & Kinetics-400 & average(Top-1 error, Top-5 error) & 12.4 \cite{bian2017revisiting} & 31 \\
          & & & TAP & ActivityNet 1.3 & AR-AN AUC & 64.80 \cite{lin2018temporal} & 17 \\
          & & & TAL/D & ActivityNet 1.3 & mAP tIoU@0.5:0.05:0.95 & 33.40 \cite{lin2018temporal} & 17 \\
          \hline
         % http://activity-net.org/challenges/2018/challenge.html
         % https://arxiv.org/abs/1808.03766
         ActivityNet \cite{ghanem2018activitynet} & 2018 & CVPR & AR & Kinetics-600 & average(Top-1 error, Top-5 error) & 10.99 \cite{he2018exploiting} & 13 \\
          & & & AR & MiT (full-track) & average(Top-1 acc, Top-5 acc) & 52.91 \cite{li2018team} & 29 \\
          & & & AR & MiT (mini-track) & average(Top-1 acc, Top-5 acc) & 47.72 \cite{guansysu} & 12 \\
          & & & TAP & ActivityNet 1.3 & AR-AN AUC & 71.0 \cite{ghanem2018activitynet} & 55 \\
          & & & TAL/D & ActivityNet 1.3 & mAP tIoU@0.5:0.05:0.95 & 38.53 \cite{Lin_2018_ECCV} & 43 \\
          & & & SAL/D & AVA & frame-mAP sIoU@0.5 & 20.99 \cite{ava2018topresult} & 23 \\
          \hline
         % http://activity-net.org/challenges/2019/challenge.html
         ActivityNet \cite{activitynetchallenge2019, Lee_2020_WACV} & 2019 & CVPR & AR & Kinetics-700 & average(Top-1 error, Top-5 error) & 17.88 \cite{qiu2019trimmed} & 15 \\
          & & & AR & EPIC-KITCHENS-55 & micro-avg Top-1,5 acc, macro-AP,AR & 41.4, 25.1 \cite{wang2019baiduuts} & 25 \\
          & & & AP & EPIC-KITCHENS-55 & micro-avg Top-1,5 acc, macro-AP,AR & 13.2, 8.5 \cite{Furnari_2019_ICCV} & 8 \\
          & & & TAP & ActivityNet 1.3 & AR-AN AUC & 72.99 & 61 \\
          & & & TAL/D & ActivityNet 1.3 & mAP tIoU@0.5:0.05:0.95 & 39.7 & 23 \\
          & & & TAL/D & VIRAT & P$_{rate}$@miss$_{FA}$ & 0.605 & 42 \\ % \cite{actev-results-2019}
          & & & SAL/D & AVA & frame mAP sIoU@0.5 & 34.25 \cite{Feichtenhofer_2019_ICCV} & 32 \\
          \hline
         % https://sites.google.com/view/multimodalvideo/home
         Multi-modal \cite{multimodalICCV19} & 2019 & ICCV & AR & Multi-MiT & mAP & 0.608 \cite{zhang2020top1} & 10 \\
          & & & TAL/D & HACS Segments & mAP tIoU@0.5:0.05:0.95 & 23.49 \cite{zhang2019learning} & 5 \\
          \hline
         % http://activity-net.org/challenges/2020/index.html
         ActivityNet \cite{activitynetchallenge2020} & 2020 & CVPR & AR & Kinetics-700 & average(Top-1 error, Top-5 error) & 14.9 & 4 \\
          & & & TAL/D & ActivityNet 1.3 & mAP tIoU@0.5:0.05:0.95 & 42.79 \cite{wang2020cbrnet} & 55 \\
          & & & TAL/D & MEVA \cite{Corona_2021_WACV} & avg(1-$P_{\text{miss}}$) across TFA from 0-20\% & 0.350 & 11 \\ % \cite{actev-results-2020}
          & & & TAL/D & HACS Segments & mAP tIoU@0.5:0.05:0.95 & 40.53 \cite{qing2020temporal} & 22 \\
          & & & TAL/D & HACS Clips+Seg. & mAP tIoU@0.5:0.05:0.95 & 39.29 \cite{gaomulti} & 13 \\
          & & & SAL/D & AVA-Kinetics & frame mAP sIoU@0.5 & 39.62 \cite{chen20201st} & 11 \\
          \hline
          % http://activity-net.org/challenges/2021/
          ActivityNet & 2021 & CVPR & AR & Kinetics-700-2020 \cite{DBLP:journals/corr/abs-2010-10864} & average(Top-1 error, Top-5 erro) & 14.0 & 27 \\
          & & & AR & TinyVIRAT \cite{demir2020tinyvirat} & F1-score & 0.478 \cite{tirupattur2021tinyaction} & 3+ \\
          & & & TAL/D & ActivityNet 1.3 & mAP tIoU@0.5:0.05:0.95 & 44.67 \cite{DBLP:journals/corr/abs-2106-11812} & 3+ \\
          & & & TAL/D & HACS Segments & mAP tIoU@0.5:0.05:0.95 & 44.29 \cite{DBLP:journals/corr/abs-2106-13014} & 22 \\
          & & & TAL/D & HACS Clips+Seg. & mAP tIoU@0.5:0.05:0.95 & 22.45 \cite{DBLP:journals/corr/abs-2106-13014} & 13 \\
          & & & TAL/D & SoccerNet-V2 \cite{deliege2021soccernetv2} & average mAP per class & 74.84 \cite{zhoukangcheng2021} & 7 \\
          & & & TAL/D & MEVA & avg(1-$P_{\text{miss}}$) across TFA from 0-20\% & 0.425 & 9 \\
          & & & SAL/D & AVA-Kinetics & frame mAP sIoU@0.5 & 40.67 \cite{DBLP:journals/corr/abs-2106-08061} & 11 \\
         \hline
         %\multicolumn{7}{l}{AR = Action Recognition} \\
         %\multicolumn{7}{l}{TAP = Temporal Action Proposal} \\
         %\multicolumn{7}{l}{TAL/D = Temporal Action Localization/Detection} \\
         %\multicolumn{7}{l}{SAL/D = Spatiotemporal Action Localization/Detection} \\
    \end{tabular}
    \label{tab:app-competitions}
\end{table*}

Several competitions introduced state-of-the-art datasets, galvanized model development, and standardized metrics.  THUMOS Challenges \cite{THUMOS13, THUMOS14, THUMOS15} were conducted through the International Conference on Computer Vision (ICCV) in 2013, the European Conference on Computer Vision (ECCV) in 2014, and the Conference on Computer Vision and Pattern Recognition (CVPR) in 2015. These primarily focused on AR and TAL/D tasks. ActivityNet Large Scale Activity Recognition Challenges \cite{Heilbron_2015_CVPR, ghanem2017activitynet, ghanem2018activitynet, activitynetchallenge2019, activitynetchallenge2020} were held at CVPR from 2016 through 2020 and have slowly expanded in scope encompassing trimmed AR, untrimmed AR, TAP, TAL/D, and SAL/D competitions.  Other challenges were modeled off THUMOS and ActivityNet such as the Workshop on Multi-modal Video Analysis and Moments in Time Challenge\footnote{\url{https://sites.google.com/view/multimodalvideo/home}} held at ICCV in 2019.  We provide an overview of these highly visible competitions in Table \ref{tab:app-competitions}. We anticipate these competitions will continue to grow in popularity in future years.  Note that the metrics specified in the table will be defined and described in Section \ref{metrics}. These competitions are a useful place for finding the current state-of-the-art models and methods. Check the "Top Results" column of Table \ref{tab:app-competitions} for papers describing the best-performing models.

\section{Data Preparation}\label{data_preparation}

While some datasets are available in pre-processed forms, others are presented \textit{raw}---using the original frame rate, frame dimensions, duration, and/or formatting.  \textit{Data preparation} is the process of transforming data prior to learning. This step is essential to extract relevant features, fit model input specifications, and prevent overfitting during training.  Key preparation processes include:
\begin{itemize}
    \item \textit{Data cleaning} is the process of detecting and removing incomplete or irrelevant portions of the dataset.  For datasets that simply link to YouTube or other web videos (e.g., \cite{kay2017kinetics, carreira2018short, carreira2019short, Zhao_2019_ICCV}), this step of determining which videos are still active on the site could be very important and affect the dataset quality.
    \item \textit{Data augmentation} is the process of transforming data to fit model input specifications and increase data diversity.  Data diversity helps prevent \textit{overfitting}---when a model too closely matches training data and fails to generalize to unseen examples.  Overfitting can occur when the model learns undesired, low-level biases rather than desired, high-level semantics.
    \item \textit{Hand-crafted feature extraction} is the process of transforming raw RGB video data into a specified feature space to provide insights that a model may not be able to independently learn.  With video data, motion representations are the most common extracted features.
\end{itemize}

While data cleaning is certainly important, this section focuses primarily on augmentation and hand-crafted feature extraction because of the many video domain-specific terms.

\subsection{Video Data}

Video is composed of a series of still-image frames where each frame is made of rows and columns of \textit{pixels}. Pixels are the smallest elements of raster images.  In standard 3-channel red-green-blue (RGB) video, each pixel is a 3-tuple with an intensity value from 0 to 255 for each of the three color channels.  RGB-D video contributes a fourth channel that represents depth, often determined by a depth sensor such as the Microsoft Kinect.\footnote{\url{https://developer.microsoft.com/en-us/windows/kinect/}}

As used throughout this article, a common abstraction to represent video is a 3-dimensional (3D) volume in which frames are densely stacked along a temporal dimension.  However, with multi-channel pixels, this volume actually has four dimensions.  The desired order of these dimensions can vary between software packages with \textit{(frames, channels, height, width)} known as channels first (NCHW) and \textit{(frames, height, width, channels)} known as channels last (NHWC).  This order can lead to performance improvements or degradation depending on the training environment (e.g., Theano\footnote{\url{https://deeplearning.net/software/theano/}} and MXNet\footnote{\url{https://mxnet.apache.org/versions/1.6/}} versus CNTK\footnote{\url{https://docs.microsoft.com/en-us/cognitive-toolkit/}} and TensorFlow\footnote{\url{https://www.tensorflow.org/}}).

\subsection{Data Augmentation}

\begin{figure*}
    \centering
    \includegraphics[width=1.0\linewidth]{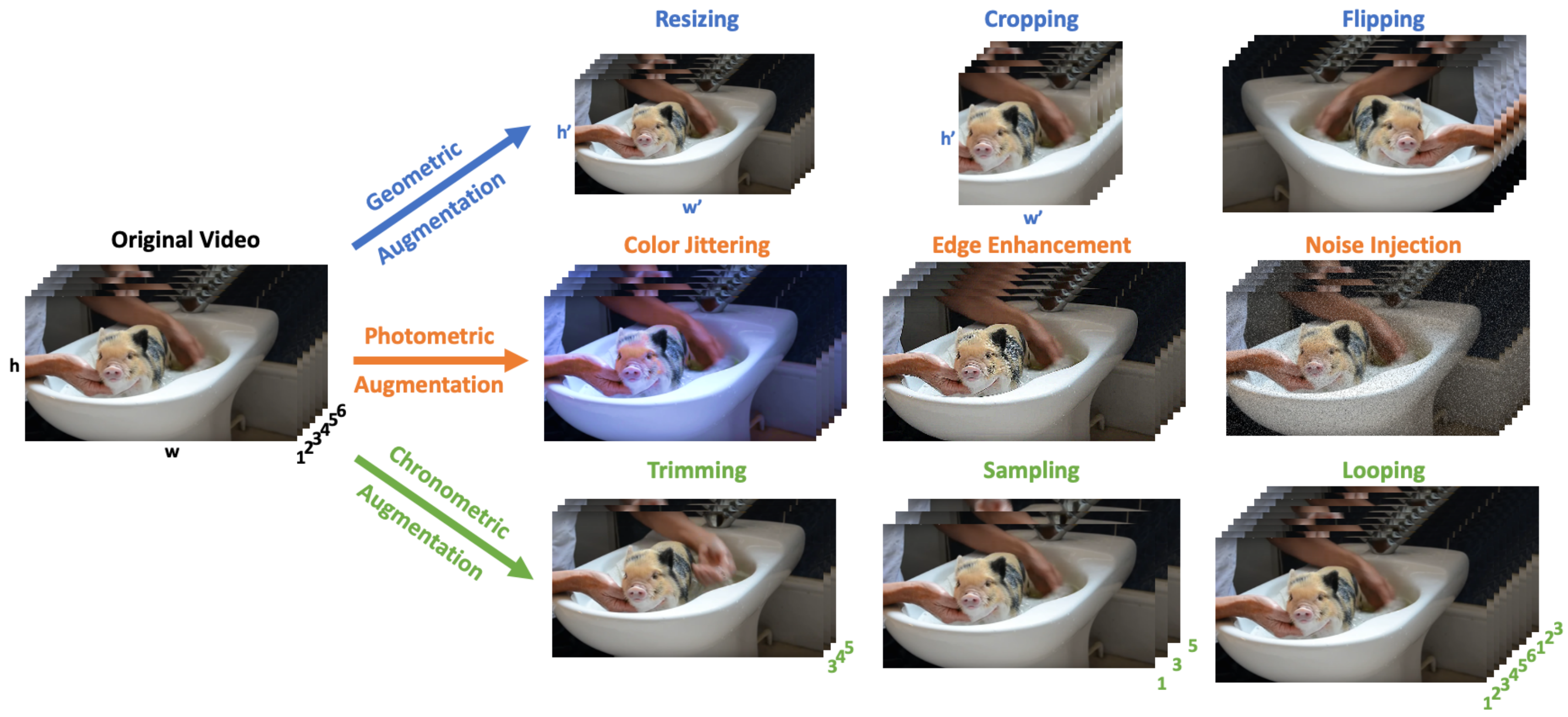}
    \caption{Common video augmentations. The top row shows three geometic augmentations that affect geometry of the frames.  These include resizing which changes the height and width while either preserving or permuting the aspect ratio, cropping which removes pixels, and horizontal flipping which mirrors across the vertical center axis. The middle row shows three photometric augmentations that affect the color-space of the frames.  These include color jittering which adjusts pixel color in an orchestrated way, edge enhancement which increases the definition of contours, and noise injection which adjusts pixel color in a random way.  The bottom row shows three chronometic augmentations that affect the temporal aspect of the video.  These include trimming which removes consecutive sections of frames, sampling which removes periodic frames, and looping which duplicates frame sequences and adds that to the beginning or end of the original frame sequence. Frames are taken from the Moments in Time dataset \cite{8651343} "washing" class.}
    \label{fig:augmentations}
\end{figure*}

\subsubsection{Geometric Augmentation Methods}

In the context of video, \textit{geometric augmentation methods} are transformations that alter the geometry of frames \cite{8628742}.  To be effective, these must be applied equally across all frames.  If separate geometric transformations are applied on different frames, a video could quickly lose its semantic meaning. Common geometric augmentations include:
\begin{itemize}
    \item \textit{Resizing}---the process of scaling a video's frames from a given height and width $(h, w)$ to a new height and width $(h',w')$ via spatial up-sampling or down-sampling \cite{imageresizing}. Ratio jittering \cite{10.1007/978-3-319-46484-8_2} is resizing that permutes the aspect ratio done for data diversification.
    \item \textit{Cropping}---the process of transforming a video's frames from a given height and width $(h,w)$ to a new, smaller height and width $(h',w')$ via removing exterior rows or columns.  Techniques include random cropping \cite{NIPS2012_4824, chatfield2014return,imageaugmentationsurvey} and corner cropping \cite{8454294}.
    \item \textit{Horizontal (left-right) flipping}---the process of mirroring a video's frames across the vertical axis (i.e., reversing the order of columns in each frame). Random horizontal flipping is a popular and computationally efficient method of introducing data diversity \cite{NIPS2012_4824, NIPS2014_5353, carreira2017quo, 8454294}.
\end{itemize}

Other geometric augmentation methods that are less popular for video include \textit{vertical flipping}, \textit{shearing}, \textit{piecewise affine transforming}, and \textit{rotating}.  Shorten and Khoshgoftaar (2019) \cite{imageaugmentationsurvey} present a survey on image augmentation which describes some of these alternative techniques that could easily be applied to video. Some might be more likely to change the semantic meaning of actions.  For example, jumping is an action generally predicated on an actor moving upward.  Vertical flipping or a 180 degree rotation would change the apparent direction of motion possibly confusing the model into believing the action is falling.

\subsubsection{Photometric Augmentation Methods}

In the context of video, \textit{photometric augmentation methods} are transformations that alter the color-space of the pixels making up each frame \cite{8628742}.  Unlike geometric augmentation, these transformations can generally be applied on a per-frame basis and are overall less common in the action understanding literature.  These include:
\begin{itemize}
    \item \textit{Color jittering}---the process of transforming a video's hue, saturation, contrast, or brightness.  This can be done randomly \cite{Wu_2015_CVPR, Han_2019_ICCV, NIPS2014_5353} or via a specific adjustments \cite{Razavian_2014_CVPR_Workshops,NIPS2012_4824}.
    \item \textit{Edge enhancement}---the process of increasing the appearance of contours in a video's frames. In some settings, this speeds up the learning process since it is shown that the first few layers in convolutional neural networks learn to detect edges and gradients \cite{NIPS2012_4824}.
\end{itemize}

Other photometric augmentation methods that may be useful in future settings are \textit{superpixelization}, \textit{random gray} \cite{Han_2019_ICCV}, \textit{random erasing} \cite{imageaugmentationsurvey}, and \textit{vignetting} \cite{Han_2019_ICCV}.  However, these are not only absent from the action understanding literature but also uncommon in image understanding.

\subsubsection{Chronometric Augmentation Methods}

Because the literature does not appear to have a term for transformations that affect the duration of the video input, we refer to these as \textit{chronometric augmentations} following the naming pattern of geometric and photometric.  These transformations are generally used to fit a model's input specifications rather than increase data diversity.
\begin{itemize}
    \item \textit{Trimming}---the process of altering the start and end of a video---essentially temporal cropping. This may be useful to remove the portion of the video that does not include the labeled action.
    \item \textit{Sampling}---the process of extracting frames from a video---essentially temporal resizing.  This can be done from specific frame indices  \cite{Feichtenhofer_2016_CVPR, carreira2017quo} or randomly selected frame indices \cite{NIPS2014_5353, 8454294}. 
    \item \textit{Looping}---the process of repeating a video's frames to increase the duration---essentially temporal padding \cite{carreira2017quo}.  This might be necessary when a video segment has fewer frames than the model's input specifies.
\end{itemize}

\subsection{Hand-Crafted Feature Extraction}

\begin{figure*}
    \centering
    \includegraphics[width=1.0\linewidth]{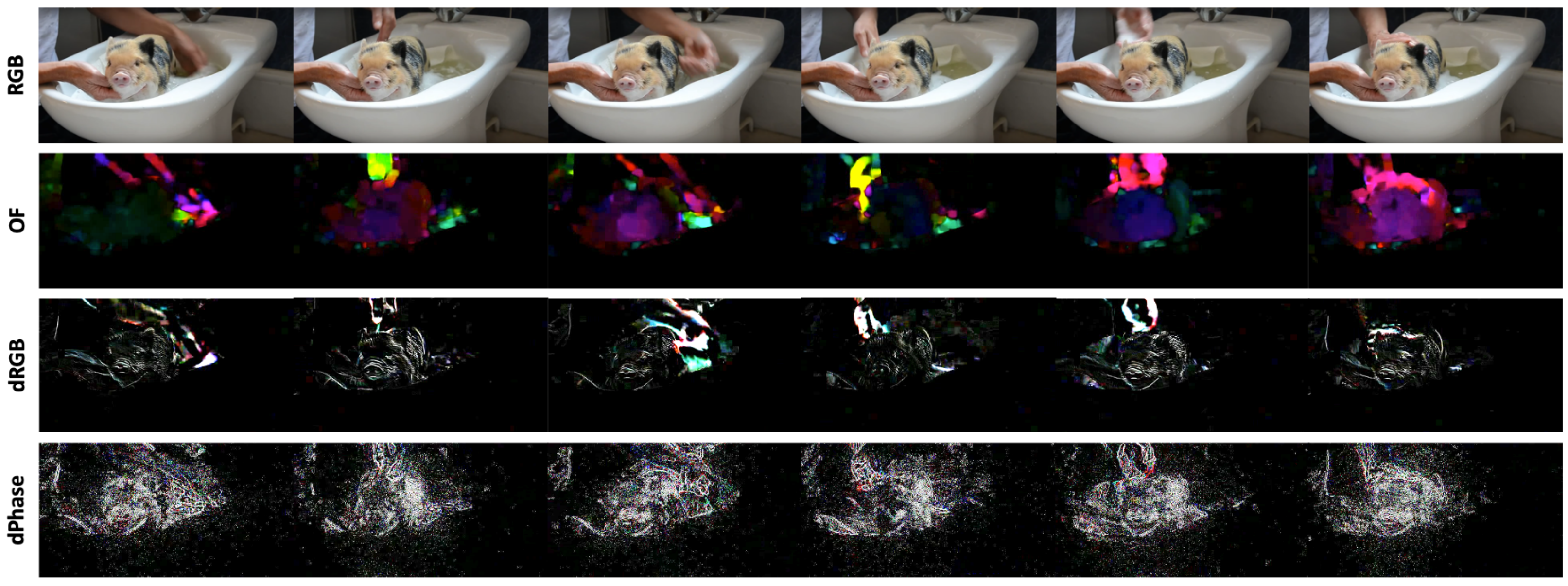}
    \caption{Examples of hand-crafted video features.  The first row shows the original RGB sampled video frames. The second row shows dense optical flow (OF) computed using the Farneback method \cite{10.1007/3-540-45103-X_50} and OpenCV packages \cite{opencv_library} (color indicates direction).  The third row shows RGB difference/derivative (dRGB) between consecutive frames.  The fourth row shows phase difference/derivative (dPhase) between consecutive frames computed using the approach described in \cite{Hommos_2018_ECCV_Workshops}.  The video frames are from the Moments in Time dataset \cite{8651343} "washing" class.}
    \label{fig:hand-crafted}
\end{figure*}

While shallow learning is less common since the deep learning revolution, several hand-crafted motion features have found their way into state-of-the-art deep learning models \cite{NIPS2014_5353, Feichtenhofer_2016_CVPR, carreira2017quo, 8454294}.  These motion representations generally fall under two classical field theories: Lagrangian flow \cite{ouellette2006quantitative} and Eulerian flow \cite{10.1145/2185520.2185561}.  Motion representations are one way of capturing temporal information for a video.  This can be important because temporal or casual learning remains a significant challenge in the deep learning field.

\subsubsection{Lagrangian Motion Representations}

Lagrangian flow fields track individual parcel or particle motion.  In the video context, this refers to tracking pixels by looking at nearby appearance information in adjacent frames to see if that pixel has moved. The most common Lagrangian motion representation is \textit{optical flow (OF)} \cite{gibson1950perception}.  Many methods exist for computing this feature: the Lucas–Kanade method \cite{lucas1981iterative}, the Horn–Schunck method \cite{10.1117/12.965761}, the TV-L$1$ approach \cite{10.1007/978-3-540-74936-3_22}, the Farneback method \cite{10.1007/3-540-45103-X_50}, and others \cite{10.1145/212094.212141}.  It is also possible to employ a WarpFlow technique \cite{Wang_2013_ICCV} to attempt to reduce background or camera motion.  This technique requires computing the \textit{homography}, a transformation between two planes, between frames. OF is noted for its usefulness in action understanding because it is invariant to appearance \cite{10.1007/978-3-030-12939-2_20}. 

\subsubsection{Eulerian Motion Representations}

Eulerian flow fields represent motion through a particular spatial location.  In the video context, this refers to determining visual information differences at a particular spatial location across frames. Two Eulerian motion representations are \textit{RGB difference/derivative (dRGB)} \cite{8454294, 10.1007/978-3-319-46484-8_2, Hommos_2018_ECCV_Workshops} and \textit{phase difference/derivative (dPhase)} \cite{Hommos_2018_ECCV_Workshops}.  RGB derivative is the difference between pixel color intensities at equivalent spatial locations in adjacent frames.  To compute this, one frame is subtracted from another.  Phase difference requires converting each frame into the frequency domain before taking the difference and converting back to the time domain.   

\section{Models}\label{models}

The past decade of action understanding research saw a paradigm shift from primarily shallow, hand-crafted approaches to deep learning where multi-layer artificial neural networks are able to learn complex non-linear relations in structured data. In this section, we describe network building blocks that are common across the diversity of action understanding models and organize a variety of models into groups based on similar underlying architectures.  For brevity, we do not cover loss functions or methods of supervising model training.  We recommend exploring these methods by examining the papers referenced throughout this section.    

\subsection{Model Building Blocks}

\subsubsection{Convolutional Neural Networks}\label{cnns}

No deep learning architecture component has impacted action understanding (and computer vision at large) greater than convolutional neural networks (CNNs), also commonly referred to as ConvNets. A CNN is primarily composed of convolutional, pooling, normalization, and fully-connected layers. For further details, a multitude of tutorials exist on utilizing standard CNN layers (e.g., \cite{lecun2013deep,le2015tutorial}).  CNNs are useful in video understanding because the sharing of weights dramatically decreases the number of trainable parameters and therefore reduces computational cost compared to fully-connected networks.  Generally, deeper models (i.e., those with more layers) outperform shallower models by increasing the \textit{receptive field}---the portion of the input that contributes to the feature---of individual neurons in the network \cite{NIPS2016_6203}.  However, deep models can suffer from problems like exploding or vanishing gradients \cite{doi:10.1142/S0218488598000094}.  

\begin{figure}
    \centering
    \includegraphics[width=1.0\linewidth]{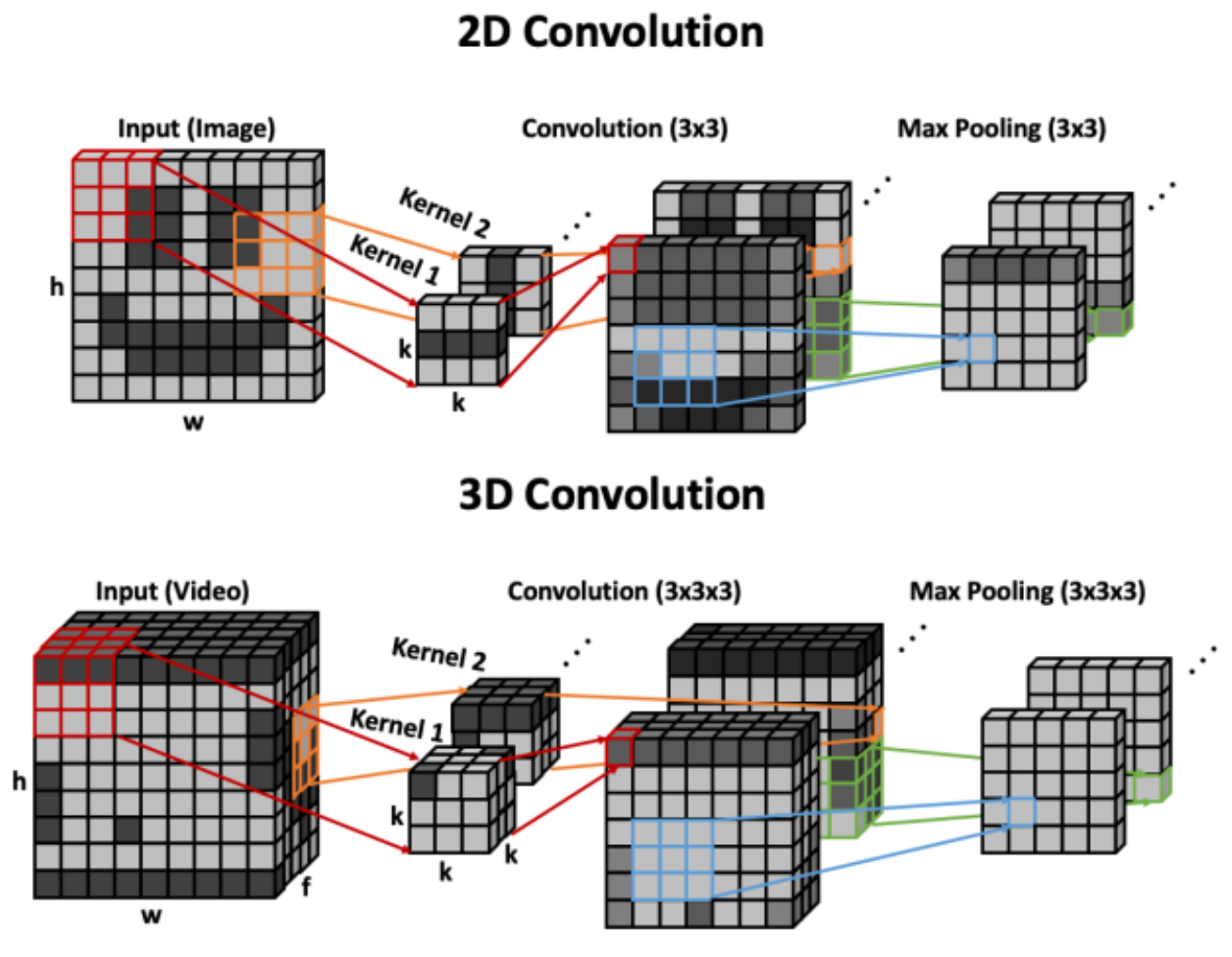}
    \caption{Examples of 2D and 3D convolutional layers and max pooling on single-channel image and video inputs. In the 2D (image) example, each small "cube" in the input represents a 3-channel pixel.  In the 3D (video) example, frames (rows and columns of pixels) are stacked from to back.  Note that these filter kernels were chosen randomly and do not necessarily lead to good embedded features.}
    \label{fig:2D3Dconv}
\end{figure}

1-Dimensional CNNs (C1D), 2-Dimensional CNNs (C2D), and 3-Dimensional CNNs (C3D) are the backbone for many state-of-the-art models and use 1D, 2D, and 3D kernels, respectively.  C1D is primarily applicable for convolutions along the time dimension of embedded features, while C2D and C3D are primarily applicable for extracting feature vectors from individual frames or stacked frames.  3D-convolution allows the for a temporal receptive field in addition to the standard spatial one.  Single-channel examples of 2D and 3D convolutions are shown in Fig. \ref{fig:2D3Dconv}.  Note that when using multi-channel inputs, the convolutional kernels must be expanded to include a depth dimension with the same number of channels as the input tensor, and the output is summed across channels.  We briefly note a few influential developments not unique to, but consistently employed throughout, the action understanding literature:
\begin{itemize}
    \item \textit{Residual networks (ResNets)} \cite{He_2016_CVPR}---utilize skip connections to avoid vanishing gradients.
    \item \textit{Inception blocks} \cite{Szegedy_2015_CVPR, Szegedy_2016_CVPR}---utilize multi-size filters for computational efficiency.
    \item \textit{Dense connections (DenseNet)} \cite{Huang_2017_CVPR}---utilize skip connections between each layer and every subsequent layer for strengthening feature propagation.
    \item \textit{Inflated networks} \cite{carreira2017quo}---expand lower dimensional networks into a higher dimension in a way that benefits from lower dimensional pretrained weights (e.g., I3D).
    \item \textit{Normalization} \cite{ioffe2015batch}---methods of suppressing the undesired effects of random initialization and random internal distribution shifts.  These include \textit{batch normalization (BN)} \cite{ioffe2015batch}, \textit{layer normalization (LN)} \cite{ba2016layer}, \textit{instance normalization (IN)} \cite{ulyanov2016instance}, and \textit{group normalization (GN)} \cite{Wu_2018_ECCV}. 
\end{itemize}

Recently, many \textit{hybrid CNNs} introduced new convolutional blocks, layers, and modules.  Some focus on reducing the large computational costs of C3D: P3D \cite{Qiu_2017_ICCV}, R(2+1)D \cite{Tran_2018_CVPR,Ghadiyaram_2019_CVPR}, ARTNet \cite{Wang_2018_CVPR}, MFNet \cite{Chen_2018_ECCV}, GST \cite{Luo_2019_ICCV}, and CSN \cite{Tran_2019_ICCV}.  Others focus on recognizing long-range temporal dependencies: LTC-CNN \cite{7940083}, NL \cite{Wang_2018_CVPR}, Timeception \cite{Hussein_2019_CVPR}, and STDA \cite{LI2020107037}.  Some unique modules include TSM \cite{Lin_2019_ICCV_TSM} which shifts individual channels along the temporal dimension for improved C2D performance, TrajectoryNet \cite{NIPS2018_7489} which uses introduces a TDD-like \cite{Wang_2015_CVPR} trajectory convolution to replace temporal convolutions, and GSM \cite{Sudhakaran_2020_CVPR} which introduces a gate-shift module.

\subsubsection{Recurrent Neural Networks}

The second most common artificial neural network architecture employed in action understanding is the \textit{recurrent neural network (RNN)}.  RNNs use a directed graph approach to process sequential inputs such as temporal data.  This makes them valuable for action understanding because frames (or frame-based extracted vectors) can be fed as inputs.  The most common type of RNN is the \textit{long short-term memory (LSTM)} \cite{doi:10.1162/neco.1997.9.8.1735}. An LSTM cell uses an input/forget/output gate structure to perform long-range learning. The second most common type of RNN is the \textit{gated recurrent unit (GRU)} \cite{cho2014learning}.  A GRU cell uses a reset/update gate structure to perform less computationally intensive learning than LSTM cells.  Several thorough tutorials cover RNN, LSTM, and GRU usage and underlying principles (e.g., \cite{chen2016gentle,grututorial, staudemeyer2019understanding, SHERSTINSKY2020132306}).

\subsubsection{Fusion}

The processes of combining input features, embedded features, or output features are known as \textit{early fusion}, \textit{middle fusion} (or \textit{slow fusion}), and \textit{late fusion} (or \textit{ensemble}), respectively \cite{Karpathy_2014_CVPR, middle-fusion, doi:10.1002/widm.1249}.  The simplest and most na\"ive form is averaging.  However, recently \textit{attention mechanisms}, processes that allow a model to focus on the most relevant information and disregard the least relevant information, have gained popularity.

\subsection{Model Architecture Families}

We focus here on grouping these methods into architecture families under each action problem and pointing to useful examples.  These lists are neither exhaustive nor intended to critique the field.  As this article is cast as a tutorial, we hope the reader gains a sense of the many varying directions of ongoing study.  Because of the rapidly evolving nature of the field, we recommend checking online scoreboards\footnote{https://paperswithcode.com/area/computer-vision} for up-to-date performances on benchmark datasets.    

\subsubsection{Action Recognition Models}\label{AR-methods}

\begin{figure*}
    \centering
    \includegraphics[width=1.0\linewidth]{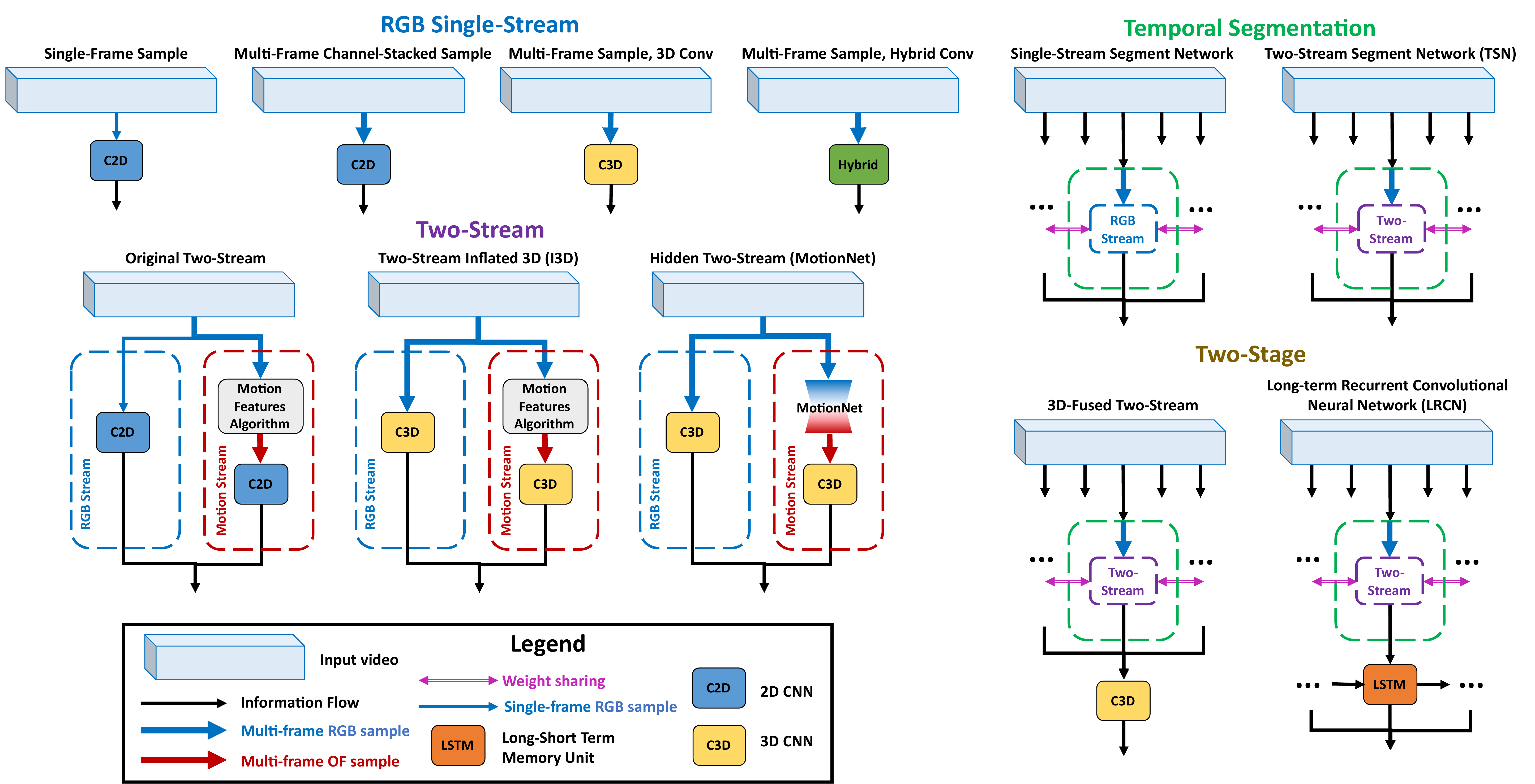}
    \caption{Action Recognition Model Examples. RGB and Motion Single-Stream architectures train a 2D, 3D, or Hybrid CNN on one sampled feature.  Two-stream architectures fuse RGB and Motion streams.  Temporal Segmentation architectures divide a video into segments, process each segment on a single-stream or multi-stream architecture, and fuse outputs.  Two-stage architectures use temporal segmentation to extract feature vectors and feed those into a convolutional or recurrent network. Please note that this is a limited selection of many action recognition model architectures.  For example, models that use recent vision transformers for video are not included above.}
    \label{fig:AR-methods-zoo}
\end{figure*}

As shown in Fig. \ref{fig:AR-methods-zoo}, we broadly group AR architectures into families of varying complexity.  The first is \textit{single-stream architectures}, which sample or extract one 2D \cite{Karpathy_2014_CVPR, he2019stnet, Jiang_2019_ICCV} or 3D \cite{10.1007/978-3-642-15567-3_11, 6165309, Tran_2015_ICCV, Hara_2018_CVPR} input feature from a video and feed that into a CNN.  The output of the CNN is the model's prediction.  While surprisingly effective at some tasks \cite{Karpathy_2014_CVPR, 8651343}, single-stream methods often lack the temporal resolution to adequately perform AR without the application of state-of-the-art hybrid modules discussed in Section \ref{cnns}.

The second family is \textit{two-stream architectures} with one stream for RGB learning and one stream for motion feature learning \cite{NIPS2014_5353, carreira2017quo}.  However, computing optical flow or other hand-crafted features is computationally expensive.  Therefore, several recent models use a "hidden" motion stream where motion representations are learned.  These include MotionNet \cite{10.1007/978-3-030-20893-6_23} which operates similarly to standard two-stream methods, and MARS \cite{Crasto_2019_CVPR} and D3D \cite{Stroud_2020_WACV} which perform middle fusion between the streams. Feichetenhofer et al. (2017) \cite{Feichtenhofer_2017_CVPR} explores gating techniques between the streams.  While these models are generally computationally constrained to two streams, more streams for additional modalities are possible \cite{WANG201733, Wang_2018}. 

Built out of single-streams, two-streams, or multi-streams, the third family is \textit{temporal segmentation architectures} which address long-term dependencies of actions.  Temporal Segment Network (TSN) methods \cite{10.1007/978-3-319-46484-8_2, 8454294} divide an input video into $N$ segments, sample from those segments, and create video-level prediction by averaging segment level outputs.  Model weights are shared between each segment stream. T-C3D \cite{liu2018t}, TRN \cite{Zhou_2018_ECCV}, ECO \cite{Zolfaghari_2018_ECCV}, and SlowFast \cite{Feichtenhofer_2019_ICCV} build on temporal segmentation by performing multi-resolution segmentation and/or fusion.  Temporal Pyramid Networks (TPN) \cite{Yang_2020_CVPR} also perform segmentation across multiple temporal levels with improved performance and robustness over standard TSN and TRN implementations.

The fourth family, a higher level of complexity, is \textit{two-stage learning} where the first stage uses temporal segmentation methods to extract segment embedded feature vectors and the second stage trains on those features.  These include 3D-fusion \cite{Feichtenhofer_2016_CVPR} and CNN+LSTM approaches \cite{Donahue_2015_CVPR, Ng_2015_CVPR, 10.1145/2733373.2806222, Wang_2019, 8904245}.  Ma et al. (2019) conducted a side-by-side comparison of C3D and CNN+LSTM performance \cite{MA201976}.  Temporal Excitation and Aggregation (TEA) \cite{Li_2020_CVPR} also uses a two-stage-like approach extracting frame-level features with a C2D prior to motion excitation and multiple temporal aggregation.  

Arguably, a fifth family of action recognition models that employs \textit{vision transformers} coalesced in 2020 and 2021. Transformers, which made their debut in the natural language processing (NLP) field in 2017 \cite{vaswani2017attention}, are an encoder-decoder sequence-to-sequence modeling schema that uses self-attention rather than recurrent neural networks or convolution.  Examples include TimeSformer \cite{DBLP:journals/corr/abs-2102-05095} which uses embeddings of frame patches augmented with positional information as a sequence of tokens for the transformer, VTN \cite{DBLP:journals/corr/abs-2102-00719} which is based off a transformer model that processes long sequences of tokens, ViViT \cite{DBLP:journals/corr/abs-2103-15691} which uses another pure-transformer architecture, and MViT \cite{DBLP:journals/corr/abs-2104-11227} which introduce resolution and channel scaling in combination with the vision transformer.

\subsubsection{Action Prediction Models}

\begin{figure*}
    \centering
    \includegraphics[width=1.0\linewidth]{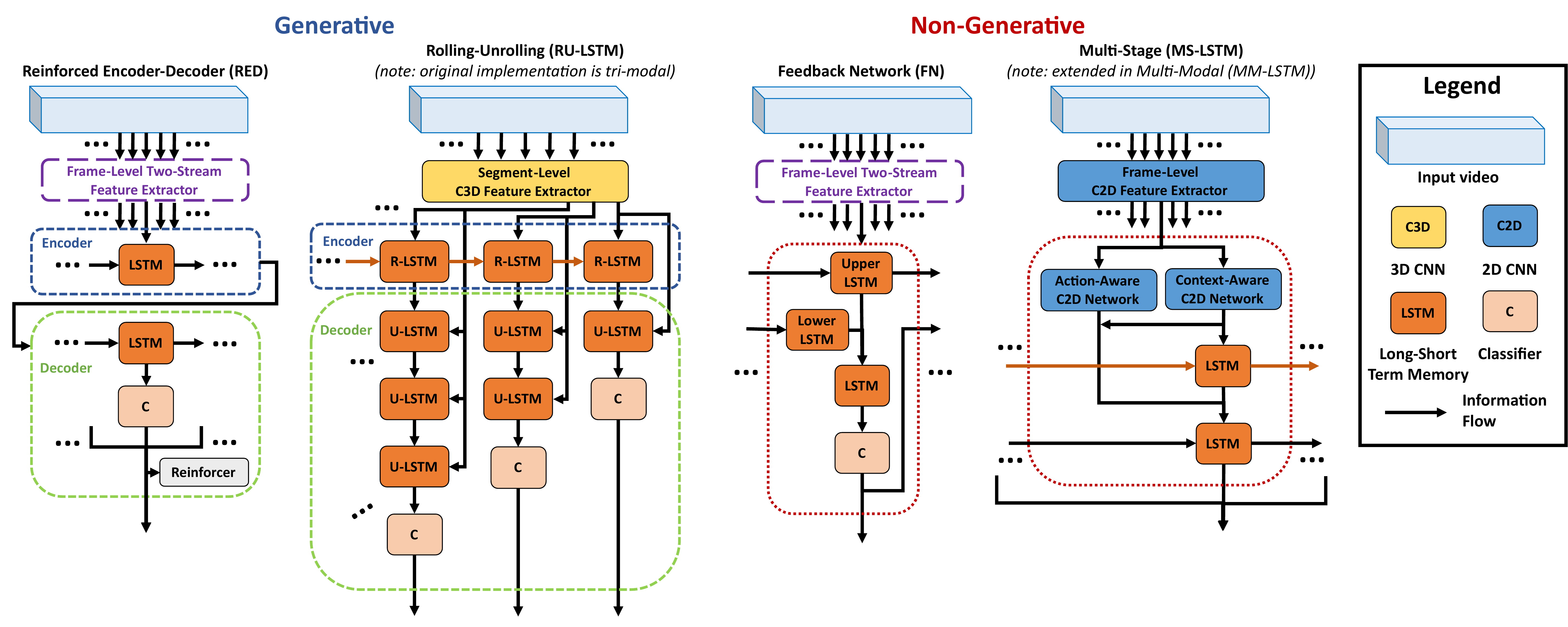}
    \caption{Action Prediction Model Examples.  Generative models create representations of future timesteps for prediction (typically via an encoder-decoder scheme). Non-generative models is a broad-sweeping category for those which create predictions directly from observed sections of the input.}
    \label{fig:AP-methods-zoo}
\end{figure*}

Rasouli (2020) \cite{rasouli2020deep} noted that recurrent techniques dominate the approaches.  We group these highly diverse action prediction models into \textit{generative} or \textit{non-generative} families.  Generative architectures produce "future" features and then classify those predictions.  This often takes the form of an encoder-decoder scheme.  Examples include RED \cite{gao2017red} which uses a reinforcement learning module to improve an encoder-decoder, IRL \cite{Zeng_2017_ICCV} which uses a C2D inverse-reinforcement learning strategy to predict future frames, Conv3D \cite{8794278} which uses a C3D to generate unseen features for prediction, RGN \cite{Zhao_2019_ICCV_RGN} which uses a recursive generation and prediction scheme with a Kalman filter during training, and RU-LSTM \cite{Furnari_2019_ICCV,8803534} which uses a multi-modal rolling-unrolling encoder-decoder with modality attention. 

Non-generative architectures is a broad grouping of all other approaches.  These create predictions directly from observed features.  Examples include F-RNN-EL \cite{7487478} which uses an exponential loss to bias a multi-modal CNN+LSTM fusion strategy towards the most recent predictions, MS-LSTM \cite{Aliakbarian_2017_ICCV} which uses two LSTM stages for action-aware and context-aware learning, MM-LSTM \cite{10.1007/978-3-030-20887-5_28} which extends MS-LSTM to arbitrarily many modalities, FN \cite{8354277} which uses a three-stage LSTM approach, and TP-LSTM \cite{8803820} which uses a temporal pyramid learning structure.

Many of these examples in this section were developed for action anticipation (when no portion of the action is yet observed), but they are also applicable for early action recognition (when a portion of the action was observed).  Additionally, action recognition models described in Section \ref{AR-methods} may be applicable for some early-action recognition tasks if they are able to derive enough semantic meaning from the provided portion and the video context.

\subsubsection{Temporal Action Proposal Models}\label{TAP-methods}

\begin{figure*}
    \centering
    \includegraphics[width=1.0\linewidth]{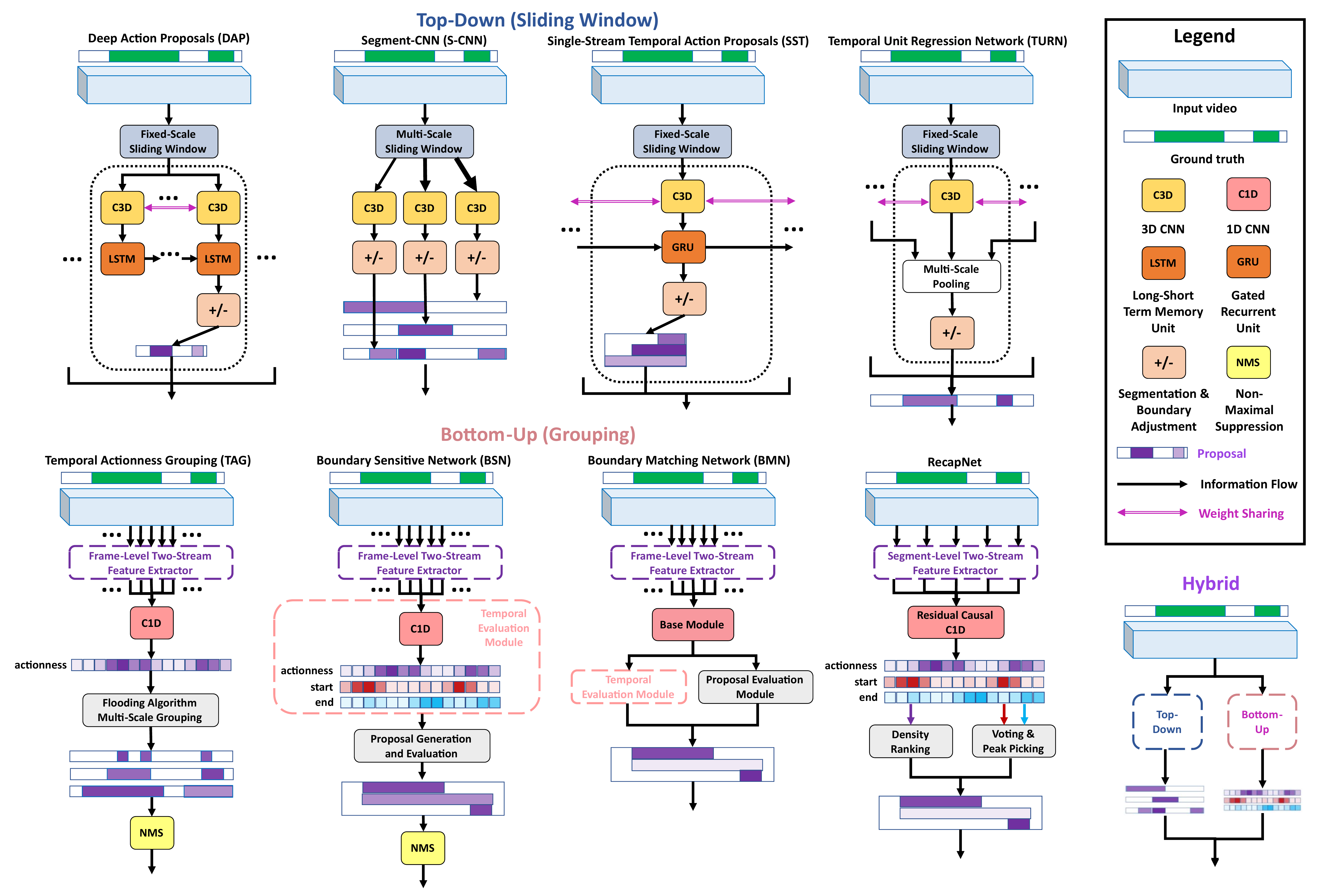}
    \caption{Temporal Action Proposal Model Examples.  Top-down models use a sliding window approach to create segment-level proposals.  Bottom-up models use frame or short-segment level actionness score predictions with grouping strategies to produce proposals.  Hybrid models use both top-down and bottom-up strategies in parallel.}
    \label{fig:TAP-methods-zoo}
\end{figure*}

As shown in Fig. \ref{fig:TAP-methods-zoo}, TAP approaches can be grouped into three families.  The first family is \textit{top-down architectures} which use sliding windows to derive segment-level proposals.  Examples include DAP \cite{10.1007/978-3-319-46487-9_47} and SST \cite{Buch_2017_CVPR} which use CNN feature extractors and recurrent networks, S-CNN \cite{Shou_2016_CVPR} which uses multi-scale sliding windows, and TURN TAP \cite{Gao_2017_ICCV} which uses a multi-scale pooling strategy.  

The second family is \textit{bottom-up-architectures} which use two-stream frame-level or short-segment-level extracted features to derive "actionness" confidence predictions. Various grouping strategies are then applied to these dense predictions to create full proposals. Examples include TAG \cite{Zhao_2017_ICCV} which uses a flooding algorithm to convert these predictions into multi-scale groupings,   BSN \cite{Lin_2018_ECCV} and BMN \cite{Lin_2019_ICCV} which use additional "startness" and "endness" features for different proposal generation and proposal evaluation techniques, and   RecapNet \cite{8972408} which uses a residual causal network rather than a generic 1D CNN to compute confidence predictions. R-C3D \cite{Xu_2017_ICCV} and TAL-Net \cite{Chao_2018_CVPR} use region-based methods to adapt 2D object proposals in images to 1D action proposals in videos.  Many of the bottom-up-architectures require non-maximal suppression (NMS) of outputs to reduce the weight of redundant proposals.

The third family is \textit{hybrid architectures} which combine top-down and bottom-up approaches.  These architectures generally create segment proposals and actionness scores in parallel and then use actionness to refine the proposals. Examples include CDC \cite{Shou_2017_CVPR}, CTAP \cite{Gao_2018_ECCV}, MGG \cite{Liu_2019_CVPR}, and DPP \cite{10.1007/978-3-030-36718-3_40}.

\subsubsection{Temporal Action Localization/Detection Models}

\begin{figure*}
    \centering
    \includegraphics[width=1.0\linewidth]{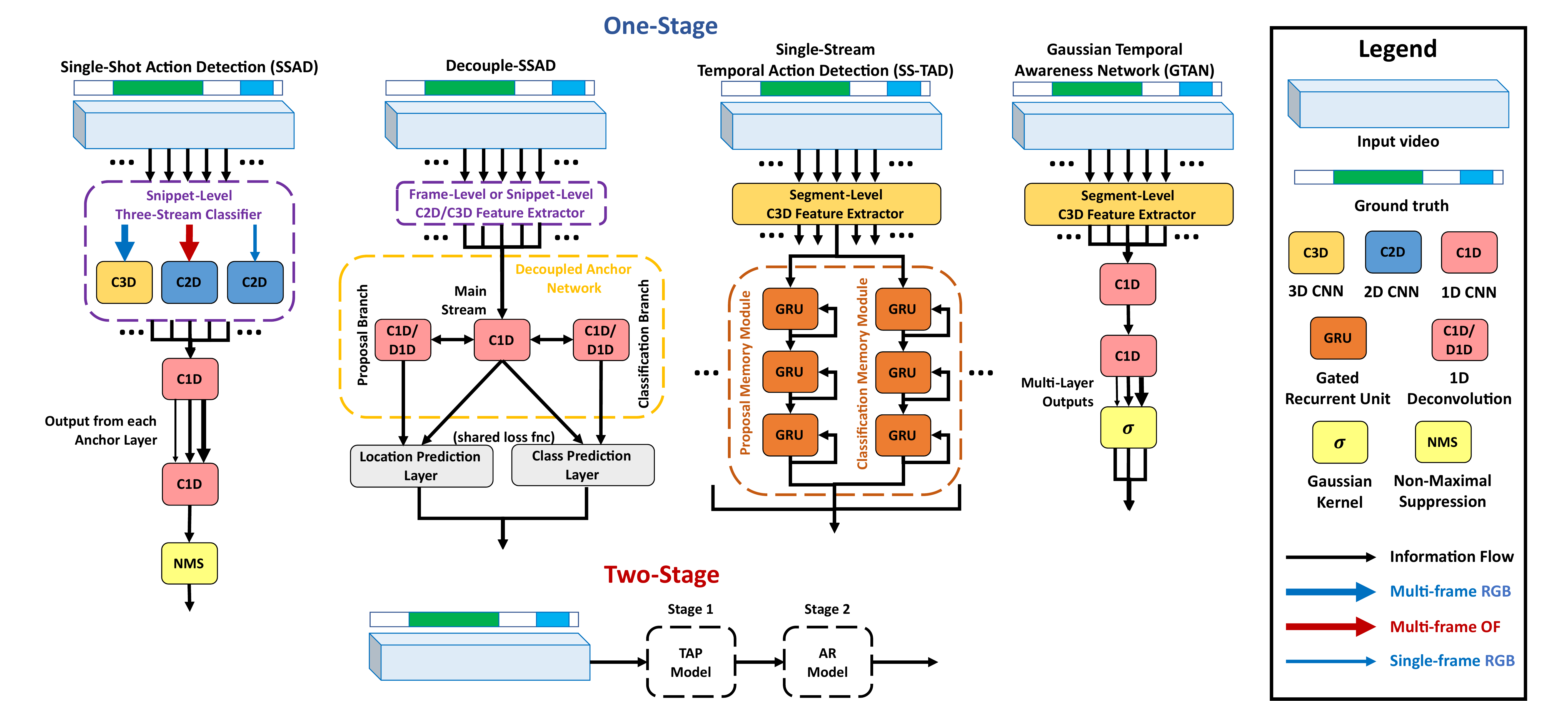}
    \caption{Temporal Action Localization/Detection Model Examples.  One-stage architectures conduct proposal and classification together while two-stage architectures create proposals and then use an action recognition model to classify each proposal.}
    \label{fig:TALD-methods-zoo}
\end{figure*}

As shown in Fig. \ref{fig:TALD-methods-zoo} and introduced in Xia et al. (2020) \cite{9062498}, there are two main families of TAL/D methods.  The first family is \textit{two-stage architectures} in which the first stage creates proposals and the second stage classifies them.  Therefore, you can pair any TAP model described in Section \ref{TAP-methods} with an AR model described in Section \ref{AR-methods}.  It is worth noting that almost all papers that explore TAP methods also extend their work to TAL/D. 

The second family is \textit{one-stage architectures} in which proposal and classification happen together.  This family of architectures spans a wide variety of implementations.  Examples include SSAD \cite{10.1145/3123266.3123343} which creates a snippet-level action score sequence from which a 1D CNN extracts multi-scale detections, SS-TAD \cite{BMVC2017_93} in which parallel recurrent memory cells create proposals and classifications, Decouple-SSAD \cite{8784822} which builds on SSAD with a three-stream decoupled-anchor network, GTAN \cite{Long_2019_CVPR} which uses multi-scale Gaussian kernels, Two-stream SSD \cite{9108686} which fuses RGB detections with OF detections, and RBC \cite{9053319} which completes boundary refinement prior to classification.  Recently, models that use graph convolutional networks have shown promise.  Examples include those proposed by Zeng et al. (2019) \cite{Zeng_2019_ICCV} and Huang et al. (2020) \cite{Huang_2020_CVPR}.

\subsubsection{Spatiotemporal Action Localization/Detection Models}

\begin{figure*}
    \centering
    \includegraphics[width=1.0\linewidth]{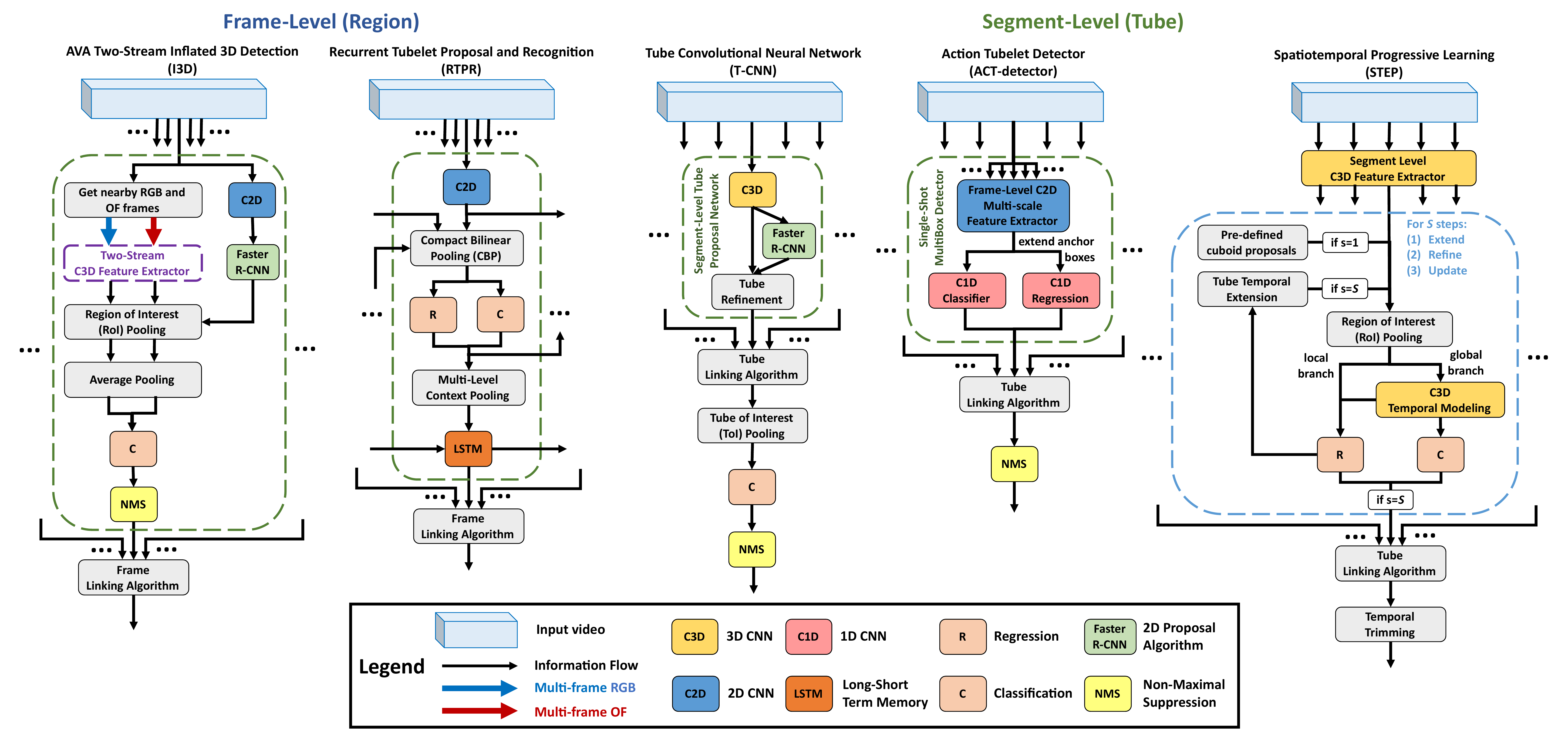}
    \caption{Spatiotemporal Action Localization/Detection Model Examples.  Frame-level (region) proposal models link frame-level detections together while segment-level (tube) proposal models create small "tubelets" for short segments and link the tubelets into longer tubes.}
    \label{fig:STALD-methods-zoo}
\end{figure*}

As shown in Fig. \ref{fig:STALD-methods-zoo}, there are two main families of state-of-the-art SAL/D methods.  The first is \textit{frame-level (region) proposal architectures} which use various region proposal algorithms (e.g., R-CNN \cite{Girshick_2014_CVPR}, Fast R-CNN \cite{Girshick_2015_ICCV}, Faster R-CNN \cite{NIPS2015_5638}, early+late fusion Faster R-CNN \cite{YE2019515}) to derive bounding boxes from frame then apply a frame linking algorithm.  Examples include MR-TS \cite{10.1007/978-3-319-46493-0_45}, CPLA \cite{yang2017spatiotemporal}, ROAD \cite{Singh_2017_ICCV}, AVA I3D \cite{Gu_2018_CVPR}, RTPR \cite{Li_2018_ECCV-RTPR}, and PntMatch \cite{YE2019515}.

The second family is \textit{segment-level (tube) proposal architectures} which uses various methods to create segment-level temporally-small tubes or "tubelets" and then uses a tube linking algorithm.  Examples of these models include T-CNN \cite{Hou_2017_ICCV}, ACT-detector \cite{Kalogeiton_2017_ICCV}, and STEP \cite{Yang_2019_CVPR}.

A few state-of-the-art models do not fit nicely in either of these families but are worth noting. Zhang et al. (2019) \cite{Zhang_2019_CVPR} use a tracking network and graph convolutional network to derive person-object detections.  VATX \cite{Girdhar_2019_CVPR} augments I3D approaches with a multi-head, multi-layer transformer.  STAGE \cite{tomei2019stage} introduces a temporal graph attention method.

\section{Metrics}\label{metrics}

Choosing the right metric is critical to evaluating a model properly. In this section, we define commonly used metrics and point to examples of their usage.  We do not cover binary classification metrics as the action datasets we cataloged overwhelmingly have more than two classes.  Note that any time we refer to an accuracy value, the error value can easily be computed as $e = 1 - a$.  To clarify terms, we use following notation across the metrics:
\begin{itemize}
    \item $X = \{x^{(1)},...,x^{(n)}\}$ as the set of $n$ input videos,
    \item $Y = \{y^{(1)},...,y^{(n)}\}$ as the set of $n$ ground truth annotations for the input videos,
    \item $M: X \to \widehat{Y}$ as a function (a.k.a. model) mapping input videos to prediction annotations,
    \item $\widehat{Y} = \{\hat{y}^{(1)},...,\hat{y}^{(n)}\}$ as the set of $n$ model outputs,
    \item $C = \{1,...,m\}$ as the set of $m$ action classes, and
    \item TP$_j: \mathbb{N} \to \{0,1\}$ as a function mapping rank in a list $L_j$ to 1 if the item at that rank is a true positive, 0 otherwise.
\end{itemize}

\begin{figure*}
    \centering
    \includegraphics[width=1.0\linewidth]{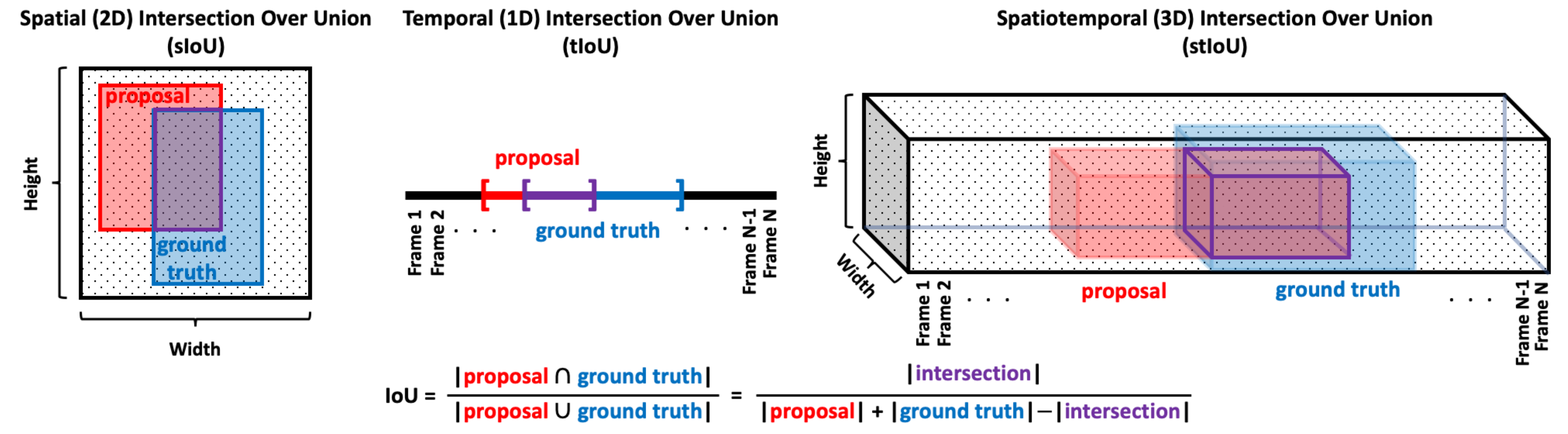}
    \caption{Illustration of types of intersection over union: spatial (left), temporal (center), and spatiotemporal (right). Below the diagrams is an equation for intersection over union (IoU).  IoU is also known as the \textit{Jaccard index} or \textit{similarity coefficient}.}
    \label{fig:IoU}
\end{figure*}

Several of these metrics also use forms of intersection over union (IoU), a measure of similarity of two regions. Fig. \ref{fig:IoU} depicts spatial IoU (sIoU), temporal IoU (tIoU), and spatiotemporal IoU (stIoU).

\subsection{Multi-class Classification Metrics}

In action understanding, multi-class classification consists of problems where the model returns per-class confidence scores for each input video. This is done primarily with a softmax loss in which the confidence scores across classes for a given input sum to one.  Formally, $\forall i \in \{1,...,n\}$:
\begin{itemize}
    \item $y^{(i)} \in C$ is the single action class ground truth annotation for input $x^{(i)}$,
    \item $\hat{y}^{(i)} = \{p^{(i)}_1,...,p^{(i)}_m \}$ where $p^{(i)}_j \in [0,1]$ is the probability that video $x^{(i)}$ depicts action $j$, and
    \item $\sum_{j=1}^{m}{p^{(i)}_j} = 1$ if the model uses softmax output (as is common).
\end{itemize}

We define the two common metrics below.  Other metrics that we do not cover include F$_1$ score (micro-averaged and macro-averaged), Cohen's Kappa, PR-AUC, ROC-AUC, partial AUC (pAUC), or two-way pAUC.  Sokolava and Lapalme \cite{SOKOLOVA2009427} and Tharwat \cite{THARWAT2018} present thorough evaluations of these and other multi-class classification metrics. 

\subsubsection{Top-$k$ Categorical Accuracy (a$_k$)}

This metric measures the proportion of times when the ground truth label can be found in the top $k$ predicted classes for that input.  Top-$1$ accuracy, sometimes simply referred to as accuracy, is the most ubiquitous while Top-$3$ and Top-$5$ are other standard choices \cite{Heilbron_2015_CVPR, ghanem2017activitynet, ghanem2018activitynet, activitynetchallenge2019, activitynetchallenge2020}.  In some cases, several Top-$k$ accuracies or errors are averaged. To calculate Top-$k$ accuracy, let $\hat{y}^{(i)}_k \subseteq \hat{y}^{(i)}$ be the subset containing the $k$ highest confidence scores for video $x^{(i)}$. The Top-$k$ accuracy ($a_k$) over the entire input set, where $\mathbbm{1}$ is a 0-1 indicator function, is defined as
\begin{equation}
    a_k = \frac{1}{n} \sum_{i=1}^{n}{\mathbbm{1}_{\hat{y}^{(i)}_k}(y^{(i)})}.
    \label{top-k}
\end{equation}

\subsubsection{Multi-Class Mean Average Precision (mAP)}\label{mAP}

This metric is the arithmetic mean of the interpolated average precision of each class, and it was used in multiple THUMOS and ActivityNet challenges \cite{THUMOS13, THUMOS14, THUMOS15, Heilbron_2015_CVPR, ghanem2017activitynet}.  To calculate interpolated average precision for a particular class $j$, the model outputs must be ranked in decreasing confidence of that class. Formally, $\forall j \in \{1,..,m\}$, $L_j$ is a ranked list of outputs such that $\forall a,b \in \{1,...,n\}, p^{(a)}_j \geq p^{(b)}_j$. The prediction at rank $r$ in list $L_j$ is a true positive if that video's ground truth label $y^{(i)}$ is class $j$ (i.e., TP$_j(r)=1$ if $y^{(i)}=j$). Using these lists $L_1$,...,$L_m$, precision (P) up to rank $k$ in a given list $L_j$ is defined as
\begin{equation}
    \text{P}_j(k) = \frac{1}{k} \sum_{r=1}^{k}{\text{TP}_j(r)}.
    \label{eqn:precision@k}
\end{equation}
Interpolated average precision (AP) over all ranks with unique recall values for a given class $j$ is calculated as
\begin{equation}
    \text{AP}(j) = \frac{ \sum_{k=1}^{n}{\text{P}_j(k) * \text{TP}_j(k) } }{ \sum_{k=1}^{n}{\text{TP}_j(k)} }.
    \label{eqn:interpolatedAP}
\end{equation}
Therefore, mean average precision (mAP) is calculated as
\begin{equation}
    \text{mAP} = \frac{1}{m} \sum_{j=1}^{m}{\text{AP}(j)}.
    \label{eqn:interpolatedmAP}
\end{equation}

\subsection{Multi-label Classification Metrics}

In the context of action understanding, multi-label classification consists of action recognition or prediction problems in which the dataset has more than two classes and each video can be annotated with multiple action class labels.  As in multi-class classification, the model returns per-class confidence scores for each input. However, in multi-label problems, it is common for the outputs to be calculated through a \textit{sigmoid loss}.  Unlike with softmax, confidence scores do not sum to one.  Formally, $\forall i \in \{1,...,n\}$:
\begin{itemize}
    \item $y^{(i)} \subseteq C$ because the ground truth annotation for input $x^{(i)}$ is a subset of action classes and
    \item $\hat{y}^{(i)}=\{p^{(i)}_1,...,p^{(i)}_m\}$ where $p^{(i)}_j \in [0,1]$ is the probability that video $x^{(i)}$ depicts action $j$.
\end{itemize}

We define two common metrics below. For more information on other metrics, such as exact match ratio and Hamming loss, we recommend Tsoumakas and Katakis (2007) \cite{tsoumakas2007multi} and Wu and Zhou (2017) \cite{pmlr-v70-wu17a} which present surveys of multi-label classification metrics.

\subsubsection{Multi-Label Mean Average Precision (mAP)}

This is the same metric as described in Section \ref{mAP}, and it is calculated very similarly for multi-label problems. The difference occurs when determining the true positives in each class list.  Here, a prediction at rank $r$ in list $L_j$ is a true positive if class $j$ is one of the video's ground truth labels (i.e., TP$_j(r)=1$ if $j \in Y^{(i)}$).  From there, precision up to rank $k$, interpolated AP for a particular class, and mAP are calculated as shown in \eqref{eqn:precision@k}, \eqref{eqn:interpolatedAP}, and \eqref{eqn:interpolatedmAP}.  This metric is used for MultiTHUMOS \cite{multithumos}, ActivityNet 1.3 \cite{Heilbron_2015_CVPR} when applied as an untrimmed action recognition problem, and Multi-Moments in Time \cite{monfort2019multimoments}.  One possible variant of multi-label mAP involves only computing AP for each class up to a specified rank $k$.  Another variant involves only counting predictions as true positives if the confidence score is above a specific threshold (e.g., $t=0.5$).

\subsubsection{Hit@k} 

This metric indicates the proportion of times when any of the ground truth labels for an input can be found in the top $k$ predicted classes for that input.  Once again, $1$ and $5$ are standard choices for $k$ \cite{Karpathy_2014_CVPR}.  Formally, let $\hat{y}^{(i)}_k \subseteq \hat{y}^{(i)}$ be the subset containing the $k$ highest confidence scores for video $x^{(i)}$.  A "hit" occurs if the intersection of the ground truth set of labels and the set of top-$k$ predictions is non-empty. Formally,
\begin{equation}
    \text{Hit@}k = \frac{1}{n} \sum_{i=1}^{n}{ [y^{(i)} \cap \hat{y}^{(i)}_k \neq \emptyset] }.
    \label{eqn:Hit@k}
\end{equation}

\subsection{Temporal Proposal Metrics}

Metrics for temporal action proposal are less varied than those for classification.  Below, we define the two main metrics found in the literature.  Here, the model returns proposed temporal regions (start and end markers for each) and a confidence score for each proposal. Formally, $\forall i \in \{1,...n\}$:
\begin{itemize}
    \item $y^{(i)} = \{s^{(i)}_1, ...\}$ is the ground truth annotation set of temporal segments where $s^{(i)}_j$ consists of start and end markers for input video $x^{(i)}$,
    \item $\hat{y}^{(i)} = \{(\hat{s}^{(i)}_1, c^{(i)}_1),...\}$ where $c^{(i)}_j \in [0,1]$ is the probability (confidence) that proposal segment $\hat{s}^{(i)}_j$ matches a ground truth segment for input $x^{(i)}$, and
    \item tIoU($s^{(i)}_a$,$\hat{s}^{(i)}_b$) is the temporal intersection over union between the ground truth and a proposal.
\end{itemize}

Intuitively, a model that produces more proposals has a better chance of covering all of the ground truth segments.  Therefore, temporal action proposal metrics include an average number of proposals (AN) hyperparameter that can be manually tuned. AN is defined as the total number of proposals divided by the total number of input videos.  Formally,
\begin{equation}
    \text{AN} = \frac{1}{n} \sum_{i=0}^{n}{|\hat{y}^{(i)}|}.
    \label{eqn:AN}
\end{equation}

\subsubsection{Average Recall at Average Number of Proposals (AR@AN)}  

Recall is a measure of sensitivity of the prediction model.  In this context, a ground truth temporal segment $s^{(i)}_a$ is counted as a true positive if there exists a proposal segment $\hat{s}^{(i)}_b$ that has a tIoU with it greater than or equal to a given threshold $t$ (i.e., TP$^{(i)}_a(t) = 1$ if tIoU$(s^{(i)}_a, \hat{s}^{(i)}_b) \geq t$).  Recall is the proportion of all ground truth temporal segments for which there is a true positive prediction. Formally, recall (R) at a particular threshold $t$ and average number of proposals (AN) is calculated as
\begin{equation}
    \text{R}(t)\text{@AN} = \frac{1}{\sum_{i=1}^{n}{|y^{(i)}|}} \sum_{i=1}^{n}\sum_{j=\{1,...\}} {\text{TP}^{(i)}_j(t)}. 
    \label{eqn:R_t@AN}
\end{equation}
Average recall is the mean of all recall values over thresholds from $0.5$ to $t_{max}$ (inclusive) with a step size of $\eta$. In the ActivityNet challenges, $t_{max} = 0.95$ and $\eta = 0.05$ \cite{ghanem2017activitynet, ghanem2018activitynet, activitynetchallenge2019}. Formally, average recall (AR) over thresholds from 0.5 to $t_\text{max}$ is calculated as
\begin{equation}
    \text{AR@AN} = \frac{1}{\frac{t_{\text{max}}-0.5}{\eta} + 1} \sum_{l=0}^{(t_{\text{max}}-0.5)/\eta}{ \text{R}(0.5+l\eta)\text{@AN} }.
    \label{eqn:AR@AN}
\end{equation}

\subsubsection{Area Under the AR-AN Curve (AUC)}

Another metric for temporal action proposal is the area under the curve when AR@AN is plotted for various values of AN.  Commonly, this is for values of 1 to 100 with a step size of 1 \cite{ghanem2017activitynet, ghanem2018activitynet, activitynetchallenge2019}. Note that at an AN of 0 where no proposals are given, AR is trivially 0.  Using AR@AN from \eqref{eqn:AR@AN}, AR-AN AUC is calculated as
\begin{equation}
    \text{AUC} = \sum_{\text{AN}=1}^{100}{ \frac{\text{AR@AN} - \text{AR@}(\text{AN}-1)}{2} }.
    \label{eqn:AUC}
\end{equation}

\subsection{Temporal Localization/Detection Metrics}

Like temporal proposal, there are two main metrics for TAL/D and both are used across many challenges \cite{THUMOS14, THUMOS15, Heilbron_2015_CVPR, ghanem2017activitynet, ghanem2018activitynet, activitynetchallenge2019, activitynetchallenge2020}. Here, the model returns proposed temporal regions (start and end markers for each), a class prediction for each proposal, and a confidence score for each proposal.  Formally, $\forall i \in \{1,...,n\}$:
\begin{itemize}
    \item $y^{(i)} = \{ (s^{(i)}_1,l^{i}_1),... \}$ is the ground truth annotation set of (temporal segment, class label) pairs for input $x^{(i)}$ where $s^{(i)}_j$ consists of start and end markers and  $l^{(i)}_j \in C$,
    \item $\hat{y}^{(i)} = \{(\hat{s}^{(i)}_1,\hat{l}^{(i)}_1,c^{(1)}_1),... \}$ where $c^{(i)}_j$ is the probability (confidence) that proposal segment $\hat{s}^{(i)}_j$ matches a ground truth segment labeled with class $\hat{l}^{(i)}_j$ for input $x^{(i)}$, and
    \item tIoU($s^{(i)}_a$,$\hat{s}^{(i)}_b$) is the temporal IoU between a ground truth segment and a proposal.
\end{itemize}

\subsubsection{Mean Average Precision at tIoU Threshold (mAP tIoU@t)}

This metric is the arithmetic mean of the interpolated average precision over all classes at a given tIoU threshold. First, all proposals for a given class are ranked in decreasing confidence. The difference from standard mAP described in Section \ref{mAP} occurs when determining true positives.  In this case, a proposal segment $\hat{s}^{(i)}_a$ at rank $r$ in list $L_j$ is counted as a true positive if there exists a ground truth segment $s^{(i)}_b$ that has a tIoU with it greater than or equal to a given threshold $t$, the predicted class label $\hat{l}^{(i)}_a$ matches the ground truth class label $l^{(i)}_b$, and that ground truth segment has not already been detected by another proposal higher in the ranked list (i.e., TP$_j(r)=1$ if tIoU$(\hat{s}^{(i)}_a,s^{(i)}_b) \geq t$ and $\hat{l}^{(i)}_a = l^{(i)}_b$).  This way, no redundant detections are allowed. Precision up to rank $k$, interpolated AP for a particular class, and mAP are calculated using \eqref{eqn:precision@k}, \eqref{eqn:interpolatedAP}, and \eqref{eqn:interpolatedmAP}.  Note that in this case, $n$ in \eqref{eqn:interpolatedAP} must be replaced with the number of prediction tuples for the class $j$.

\subsubsection{Average Mean Average Precision (average mAP)}

The most common TAL/D metric is the arithmetic mean of mAP over multiple different tIoU thresholds from $0.5$ to $t_{\text{max}}$ with a given step size $\eta$.  Commonly, $t_{\text{max}}=0.95$ (inclusive) and $\eta=0.05$ \cite{Heilbron_2015_CVPR, ghanem2017activitynet, ghanem2018activitynet, activitynetchallenge2019, activitynetchallenge2020}.  Therefore, average mAP is computed as
\begin{equation}
\begin{multlined}
    \text{average mAP} = \text{mAP tIoU@0.5:}t_{\text{max}}\text{:}\eta \\ = \frac{1}{\frac{t_{\text{max}}-0.5}{\eta} + 1} \sum_{j=0}^{(t_{\text{max}}-0.5)/\eta}{\text{mAP tIoU@}(0.5+j\eta)}.
    \label{eqn:average mAP}
\end{multlined}
\end{equation}

\subsection{Spatiotemporal Localization/Detection Metrics}

SAL/D involves locating actions in both time and space as well as classifying the located actions. Here, the model generally returns frame-level proposed spatial regions (bounding boxes), a class prediction for each box, and a confidence score. Formally, $\forall i \in \{1,...,n\}$:
\begin{itemize}
    \item $y^{(i)}=\{(f^{(i)}_1,b^{(i)}_1,l^{(i)}_1),...\}$ is the ground truth annotation set of tuples for input $x^{(i)}$ where $f^{(i)}_j$ is the frame number counting up from 1, $b^{(i)}_j$ is a bounding box marking the upper left corner, the box's height, and the box's width, and $l^{(i)}_j \in C$,
    \item $\hat{y}^{(i)}=\{(\hat{f}^{(i)}_1,\hat{b}^{(i)}_1,\hat{l}^{(i)}_1, c^{(i)}_1),...\}$ where $c^{(i)}_j$ is the confidence that bounding box $\hat{b}^{(i)}_j$ at frame $\hat{f}^{(i)}_j$ matches a ground truth bounding box on the same frame labeled with class $\hat{l}^{(i)}_j$,
    \item tube$^{(i)}_j$ is a spatiotemporal tube in video $x^{(i)}$ defined as a linked set of bounding boxes $b^{(i)}_k, b^{(i)}_l, b^{(i)}_m, ...$ with the same class label ($l^{(i)}_k = l^{(i)}_l = l^{(i)}_m = ...$) in adjacent frames ($k = l - 1 = m - 2 = ...$),
    \item sIoU$(b^{(i)}_a,\hat{b}^{(i)}_b)$ is the spatial IoU between a ground truth bounding box and a proposed bounding box (note: this requires $f^{(i)}_a=\hat{f}^{(i)}_b$), and
    \item stIoU$(\text{tube}^{(i)}_a, \widehat{\text{tube}}^{(i)}_b)$ is the spatiotemporal IoU between a ground truth and proposed tubes.
\end{itemize}

\subsubsection{Frame-Level Mean Average Precision (frame-mAP)}

This metric is useful because it evaluates the model independent of the linking strategy---the process of developing action instance tubes.  It is utilized in several ActivityNet challenges \cite{ghanem2018activitynet, activitynetchallenge2019, activitynetchallenge2020}.  Like several metrics above, this is the mean of the interpolated AP over all classes.  For a given class, every prediction tuple is ranked in decreasing confidence. Here, a proposal box $\hat{b}^{(i)}_a$ at rank $r$ in list $L_j$ is counted as a true positive if a ground truth box $b^{(i)}_b$ exists on the same frame with the same class label that has a sIoU greater than or equal to a given threshold $t$ that was not already detected by another proposed box higher in the ranked list (i.e., TP$_j(r)=1$ if sIoU$(\hat{b}^{(i)}_a,b^{(i)}_b) \geq t$ and $\hat{f}^{(i)}_a=f^{(i)}_b$ and $\hat{l}^{(i)}_a=l^{(i)}_b$). No redundant detections are allowed. Precision up to rank $k$, interpolated AP for a particular class, and mAP are calculated using \eqref{eqn:precision@k}, \eqref{eqn:interpolatedAP} and \eqref{eqn:interpolatedmAP}.  Note that $n$ in \eqref{eqn:interpolatedAP} must be replaced with the number of prediction tuples for the class $j$ (i.e., the length of ranked list $L_j$).

\subsubsection{Video-Level Mean Average Precision (video-mAP)}

This metric is useful for evaluating the linking strategy applied to connect bounding boxes of the same class label in adjacent frames.  When using frame-mAP, longer actions take up more frames and weigh more when calculating AP and mAP. However, using this metric, each action instance is weighted equally regardless of the temporal duration of the occurrence. This video-mAP metric was employed for use with both AVA and J-HMDB-21 datasets \cite{Gu_2018_CVPR, Jhuang_2013_ICCV}.  Once bounding boxes of the same class label in adjacent frames are linked into tubes, every prediction tube of that class is ranked in decreasing confidence.  Here, a proposal $\widehat{\text{tube}}^{(i)}_a$ at rank $r$ in list $L_j$ is counted as a true positive if a ground truth tube $\text{tube}^{(i)}_b$ exists with the same class label that has a stIoU greater than or equal to a given threshold $t$ that was not already detected by another proposed tube higher in the ranked list (i.e., TP$_j(r)=1$ if stIoU$(\widehat{\text{tube}}^{(i)}_a,\text{tube}^{(i)}_b) \geq t$ and $\hat{l}^{(i)}_a = l^{(i)}_b$). No redundant detections are allowed. Precision up to rank $k$, interpolated AP for a particular class, and mAP are calculated using \eqref{eqn:precision@k}, \eqref{eqn:interpolatedAP} and \eqref{eqn:interpolatedmAP}.  Note that in this case, $n$ in \eqref{eqn:interpolatedAP} must be replaced with the number of prediction tubes for the class $j$.

\section{Conclusion}\label{conclusion}

In this tutorial, we presented the suite of problems encapsulated within action understanding, listed datasets useful as benchmarks and pretraining sources, described data preparation steps and strategies, organized deep learning model building blocks and model architecture families, and defined common metrics for assessing models. We hope that this tutorial clarifies terminology and expands your understanding of these problems at the intersection of computer vision and deep learning. This article has also demonstrated the similarities and differences between these action understanding problem spaces via common datasets, model building blocks, and metrics. To that end, we also hope that this can facilitate idea cross-pollination between the somewhat stove-piped action problem sub-fields.

\section*{Acknowledgment}

We would like to thank the following individuals who have provided feedback on this article: Jeremy Kepner, Andrew Kirby, Alex Knapp, Alison Louthain, Judy Marchese, and Albert Reuther. Research was sponsored by the United States Air Force Research Laboratory and the United States Air Force Artificial Intelligence Accelerator and was accomplished under Cooperative Agreement Number FA8750-19-2-1000. The views and conclusions contained in this document are those of the authors and should not be interpreted as representing the official policies, either expressed or implied, of the United States Air Force or the U.S. Government. The U.S. Government is authorized to reproduce and distribute reprints for Government purposes notwithstanding any copyright notation herein.

\bibliographystyle{IEEEtran}
%\bibliography{refs}
\bibliography{tutorial-postedits.bbl}

\begin{IEEEbiography}[{\includegraphics[width=1in,height=1.25in,clip,keepaspectratio]{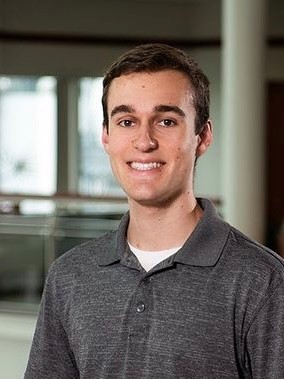}}]{Matthew S. Hutchinson} received S.B. and M.Eng. degrees in computer science and engineering from the Massachusetts Institute of Technology, Cambridge, Massachusetts, USA in 2020.  

From 2019 to 2020, he was a Graduate Research Assistant with the Massachusetts Institute of Technology (MIT) Lincoln Laboratory Supercomputing Center (LLSC).  In 2020, he commissioned into the U.S. Air Force (USAF) as a Developmental Engineer and has subsequently transferred into the U.S. Space Force (USSF).  His research interests include artificial intelligence, machine learning, computer vision.
\end{IEEEbiography}

\begin{IEEEbiography}[{\includegraphics[width=1in,height=1.25in,clip,keepaspectratio]{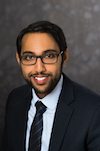}}]{Vijay N. Gadepally} received the B.Tech. degree in electrical engineering from the Indian Institute of Technology, Kanpur, India and the M.Sc. and Ph.D. degrees in electrical and computer engineering from The Ohio State University, Ohio, USA.  He is a Senior Member of the IEEE.

He is currently a Senior Member of the Technical Staff at the Massachusetts Institute of Technology (MIT) Lincoln Laboratory and leads the research activities of the Lincoln Laboratory Supercomputing Center. Additionally, he works closely with the Computer Science and Artificial Intelligence Laboratory (CSAIL) at MIT. His research interests are in high performance computing, machine learning, artificial intelligence and high-performance databases.

Dr. Gadepally's awards and honors include an Outstanding Graduate Student Award at The Ohio State University in 2011 and the MIT Lincoln Laboratory’s Early Career Technical Achievement Award in 2016. In 2017, he was named to AFCEA's inaugural 40 under 40 list.

\end{IEEEbiography}

\EOD

\end{document}